\documentclass[twocolumn]{bytedance_seed}

\usepackage{graphicx}
\usepackage{natbib}
\usepackage{caption}
\usepackage{algorithm}
\usepackage{algpseudocode}
\usepackage{booktabs}
\usepackage{multirow}
\usepackage{colortbl}
\usepackage{xcolor}
\usepackage{makecell}
\usepackage{pifont}
\usepackage{array}
\usepackage{siunitx}
\usepackage{adjustbox}
\usepackage{amsmath}
\usepackage{amssymb}
\usepackage{newfloat}
\usepackage{listings}
\usepackage[toc,page,header]{appendix}
\usepackage{xurl}
\definecolor{challengeblue}{HTML}{315F8C}
\definecolor{challengebg}{HTML}{EAF2F8}
\definecolor{baselinepink}{RGB}{239,148,158}
\definecolor{mrflowblue}{RGB}{87,127,166}
\definecolor{insetred}{RGB}{255,0,0}

\newcommand{\ReachSubet}{\mathcal Z_{\mathrm{lat}}}
\newcommand{\cmark}{\ding{51}}
\newcommand{\xmark}{\ding{55}}
\definecolor{AlgGray}{RGB}{115,115,115}
\algrenewcommand\algorithmicrequire{\textbf{Require:}}
\algrenewcommand\algorithmicensure{\textbf{Output:}}
\newcommand{\AlgPhase}[1]{\Statex {\color{blue}\normalfont\itshape // #1}}
\algrenewcommand\algorithmiccomment[1]{%
  \hfill{\color{AlgGray}\footnotesize\itshape\(\triangleright\)\ #1}}
\newcommand{\EqNote}[1]{\Comment{#1}}

\title{SelfLift: Accelerating Few-Step Diffusion via Self-Recovering Resolution Transition}

\author[1,*]{Tingyan Wen}
\author[2,*]{Chenqian Yan}
\author[2]{Xurui Peng}
\author[2]{Xiazhang Fang}
\author[2]{Shuai Wang}
\author[1,\dagger]{Xueqian Wang}
\author[2,\dagger]{Songwei Liu}

\affiliation[1]{Tsinghua University}
\affiliation[2]{ByteDance}

\contribution[*]{Equal contribution}
\contribution[\dagger]{Corresponding authors}

\abstract{
Few-step diffusion models substantially compress temporal computation, making the spatial cost of each model evaluation an increasingly dominant source of inference latency.
Progressive-resolution inference reduces this cost by performing early denoising at low resolution and reserving high-resolution computation for refinement. However, existing methods typically lift intermediate latents directly and rely on subsequent steps to absorb the induced distribution mismatch. In the few-step regime, the limited recovery budget leaves these errors as visible artifacts, constraining how late the transition can occur and, consequently, how efficiently it can be performed.
We introduce~\textbf{SelfLift}, a self-recovering progressive-resolution framework that derives both transition-repair signals and trajectory-aligned supervision from the generative model itself.
\emph{\textbf{SelfLift-zero}} proposes a training-free Artifact-Aware Consistency Lift, using disagreement between direct latent lifting and pixel-VAE re-encoding as both a localized artifact-risk signal and a model-native correction direction. It enables reliable late transitions without external super-resolution, extra denoiser evaluations, or sampling-schedule modifications.
Building on this robust transition, \emph{\textbf{SelfLift-rich}} performs On-Policy Self Recovery on student-visited states, transferring dense high-resolution guidance from an internal self-teacher while remaining aligned with the altered progressive-resolution dynamics.
Across FLUX.2-Klein and Z-Image-Turbo, SelfLift reduces end-to-end latency by \textbf{41.5\%} and \textbf{44.1\%}, respectively. Combined with timestep distillation, it delivers overall speedups of $\textbf{29.61}\times$ and $\textbf{19.21}\times$ over the corresponding 50-step models while preserving competitive generation quality, establishing a stronger speed--quality frontier for few-step diffusion.
\par\vspace{1.5em}
\noindent{\small
{\checkdatafont\sffamily\bfseries Project Page:}
\url{https://happygirlty.github.io/SelfLift_res/}\par
{\checkdatafont\sffamily\bfseries SGLang Integration (under review):}
\url{https://github.com/sgl-project/sglang/pull/33733}\par
{\checkdatafont\sffamily\bfseries Correspondence:}
Songwei Liu at \email{liusongwei.zju@bytedance.com}\par
{\checkdatafont\sffamily\bfseries Date:}
August 2026
}
}
\begin{document}
\maketitle

\begin{figure*}[t]
    \centering
    \includegraphics[width=\textwidth]{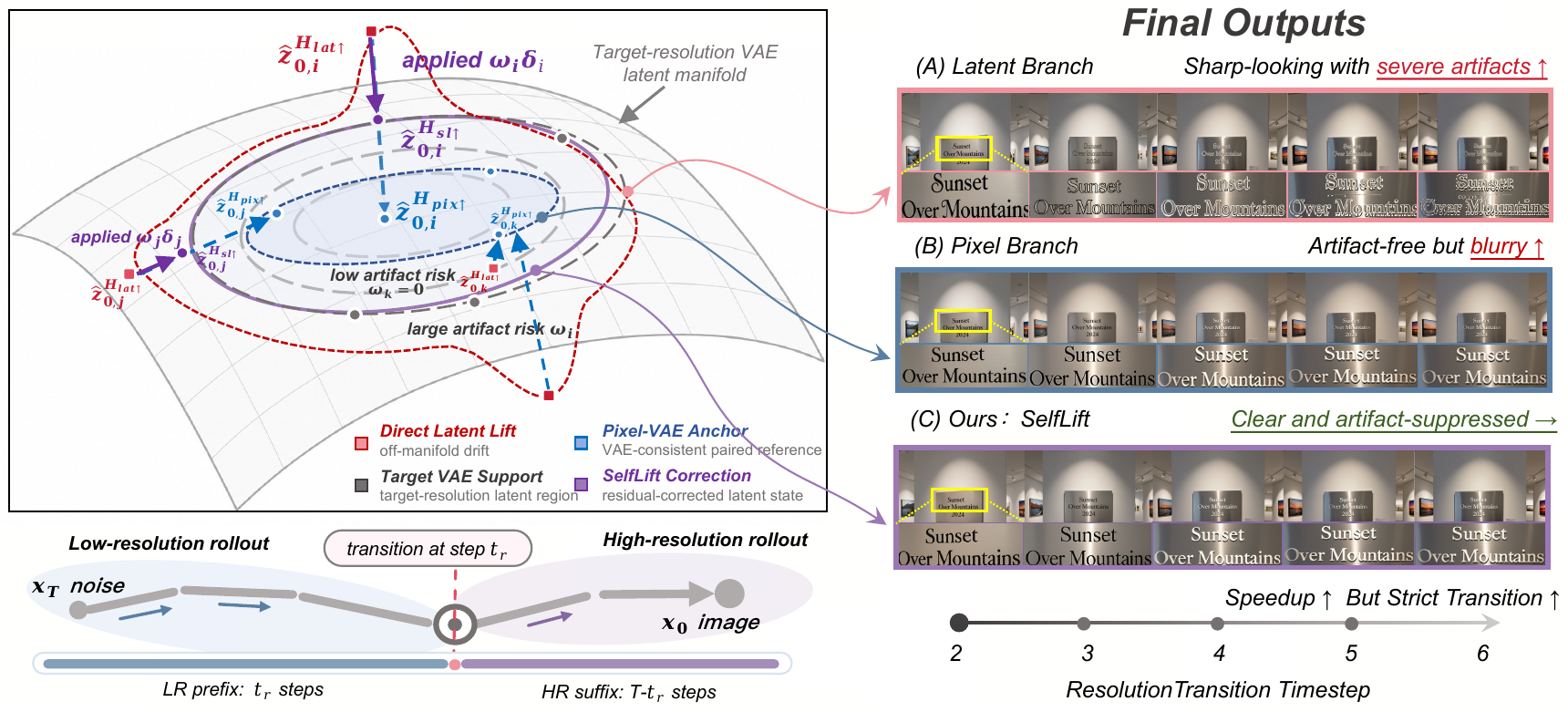}
    \caption{
    \textbf{SelfLift unlocks fast and reliable progressive-resolution inference for few-step diffusion.}
    Prior methods either leave latent-lifting artifacts to the few remaining denoising steps or rely on an external super-resolution model. SelfLift instead leverages two model-native branches, enabling later transitions with greater speedups.
    }
    \label{fig:teaser}
\end{figure*}

\section{Introduction}

Modern latent diffusion architectures achieve strong visual synthesis through iterative refinement~\cite{rombach2022latent,peebles2023dit}, but repeated full-resolution evaluations remain expensive. Step distillation compresses this process to one to eight evaluations using trajectory- or consistency- based objectives~\cite{salimans2022progressive,song2023consistency,liu2024instaflow}, distribution matching~\cite{yin2024dmd}, or variational score distillation~\cite{nguyen2024swiftbrush}. 
With temporal redundancy largely exhausted, further acceleration hinges on reducing per-step spatial cost without disrupting the distilled sparse trajectory.
Progressive-resolution inference offers a natural solution. It performs early low-resolution evaluations to establish global structure, while reserving high-resolution computation for detail refinement, following diffusion's coarse-to-fine evolution~\cite{qian2024masf,du2025postdiff}.
As the simplest and lowest-overhead bridge between resolutions, direct latent lifting is widely adopted in training-free methods~\cite{tian2025bottleneck,jeong2025ralu,xiao2026spectral}. Yet this efficiency comes at a cost, as the lifted states may exhibit visible artifacts and spatial drift, leaving the remaining high-resolution steps to recover from the resulting transition errors.

Fully exploiting progressive-resolution inference in the few-step regime poses two coupled challenges.
\textbf{Challenge 1:  Reliable resolution transition under a limited recovery budget.}
An earlier transition leaves more steps for recovery but sacrifices acceleration, whereas a later transition yields greater speedups while making these errors harder to correct. Pixel-space alternatives~\cite{zheng2026mrflow} avoid direct latent lifting through a separate super-resolution model, introducing an external generative prior and non-negligible inference overhead. Few-step acceleration therefore requires immediate, model-native transition repair without external models or extra denoiser evaluations, shifting the focus from when or where to transition to how to cross the boundary reliably.
\textbf{Challenge 2: Dense high-resolution guidance on the progressive-resolution trajectory.} A reliable training-free transition already enables stable and efficient generation. When lightweight adaptation is allowed, the model's native high-resolution capability can be further used to recover details beyond what the low-resolution prefix can represent. Standard fine-tuning and full-resolution distillation are poorly matched to this setting: they supervise target-derived or full-resolution states rather than the student-visited states produced by the accelerated rollout, creating a train–inference mismatch and potentially disrupting the few-step dynamics~\cite{miao2025pso,jiang2026dopsd}. Effective adaptation must therefore transfer dense high-resolution knowledge directly along the actual progressive-resolution trajectory used at inference.

To address these challenges, we propose \textbf{SelfLift}, a self-recovering resolution-transition framework for accelerating few-step models, in which the model itself provides both self-derived transition-repair signals and trajectory-aligned supervision. SelfLift includes a complete training-free solution, SelfLift-zero, and an optional lightweight enhancement, SelfLift-rich.
\emph{SelfLift-zero} introduces Artifact-Aware Consistency Lift. 
Specifically, it constructs a self-derived consistency residual from the discrepancy between direct latent lifting and pixel VAE re-encoding. This residual both localizes regions at risk of transition artifacts and provides a principled correction direction for the directly lifted latent. 
It can be integrated without an external super-resolution model or additional denoiser evaluations.
\emph{SelfLift-rich} further extends this principle through On-Policy Self Recovery. It distills our Zero transition into a lightweight latent lifter and transfers dense high-resolution guidance directly along the progressive-resolution trajectory. The supervision is provided by an internal self-teacher on the same student-visited states, without a separately trained teacher or reward model. 

Our contributions are summarized as follows:
\begin{itemize}
    \item We propose \textbf{SelfLift}, a self-recovering progressive-resolution framework that reframes few-step acceleration around reliable resolution transition.
    \item SelfLift introduces \textbf{two core designs}: Artifact-Aware Consistency Lift for immediate and reliable transition without external super-resolution or extra denoising, and On-Policy Self Recovery for trajectory-aligned adaptation with dense high-resolution supervision.

    \item On FLUX.2-Klein and Z-Image-Turbo, SelfLift reduces end-to-end latency by \textbf{41.5\%} and \textbf{44.1\%}. With timestep distillation, it achieves $\textbf{29.61}\times$ and $\textbf{19.21}\times$ overall speedups with strong generation quality.

\end{itemize}

\section{Related Work}
\subsection{Efficient Diffusion Inference}

Temporal acceleration employs fast ODE solvers~\cite{lu2022dpmsolver,zhao2023unipc}, error-rectified~\cite{peng2026ertacache}, multi-level~\cite{wen2025xslim}, or motion-aware caching~\cite{xu2026motioncache}, and applies feature forecasting~\cite{liu2025taylorseer}. Trajectory distillation further compresses sampling through progressive compression~\cite{salimans2022progressive}, cross-noise consistency~\cite{song2023consistency}, or distribution matching~\cite{yin2024dmd2,li2026onexdistill}. With temporal redundancy compressed in few-step models, per-evaluation spatial cost dominates.
Spatial methods reduce this cost by varying resolution along the sampling trajectory. Training-dependent designs use linked pyramids~\cite{jin2025pyramidal}, noise-dependent patchification~\cite{li2026ppflow}, or cascaded pixel-space flows~\cite{chen2025pixelflow}, but require retraining. Training-free methods include high--low--high sampling~\cite{tian2025bottleneck}, region-adaptive lifting~\cite{jeong2025ralu}, and spectrally scheduled resolution growth~\cite{xiao2026spectral}. 
MrFlow instead uses pixel-space super-resolution, VAE re-encoding, and high-resolution refinement, avoiding direct latent lifting but requiring an external super-resolution model~\cite{zheng2026mrflow}. These approaches complement temporal acceleration, yet few-step models leave little denoising budget to absorb transition errors, making a reliable and self-contained resolution transition essential.

\subsection{Post-Training of Step-distilled Models}

Step-distilled models acquire specialized dynamics over sparse timesteps, which standard post-training objectives may disrupt and thereby degrade generation quality~\cite{miao2025pso}. Effective adaptation must therefore improve the output distribution without deviating from the pretrained trajectory. PSO enlarges the likelihood margin between real images and distilled-model samples~\cite{miao2025pso}, whereas on-policy distillation supervises student-visited states to reduce the train--inference mismatch of off-policy teacher data~\cite{agarwal2024onpolicy}. Recent works extend this paradigm to visual generation~\cite{fang2026flowopd,li2026diffusionopd,jiang2026dopsd}.
These methods primarily consider post-training along native-resolution trajectories. Resolution-transition adaptation instead operates on states jointly induced by the low-resolution prefix and cross-resolution transition. It must correct the resulting distribution shift and recover high-resolution content on the actual progressive-resolution rollout, while preserving the pretrained few-step dynamics.

\section{Method}
\subsection{Preliminaries}

\paragraph{\textbf{Latent Flow Matching.}}
Given an image $x_0$, the VAE encoder $\mathcal E$ maps it to the clean latent
$z_0=\mathcal E(x_0)$. For flow time $t\in[0,1]$ and text condition $c$,
rectified flow~\cite{lipman2023flow} connects the clean latent to Gaussian noise and learns the
corresponding conditional velocity field:
\begin{equation}
\begin{gathered}
z_t=(1-t)z_0+t\epsilon,
\qquad
\epsilon\sim\mathcal N(\mathbf 0,\mathbf I),
\\
v_\theta(z_t,t,c)
\approx
\mathbb E\!\left[\epsilon-z_0\mid z_t,c\right].
\end{gathered}
\label{eq:flow_matching}
\end{equation}
Sampling follows the learned flow from $t=1$ to $t=0$ in the VAE latent
space.

\paragraph{\textbf{Progressive-Resolution Inference.}}
Let $\Phi_{a\rightarrow b}^{R}$ denote the numerical flow from time $a$ to $b$ at resolution $R$. 
Given a transition time $t_r$, progressive-resolution
inference first follows the low-resolution flow, lifts the intermediate state
to the high-resolution latent space, and then completes the remaining
trajectory at high resolution:
\begin{equation}
z_0^{\mathrm{PR}}
=
\Phi_{t_r\rightarrow 0}^{H}
\left(
\mathcal{T}_{L\rightarrow H}
\left(
\Phi_{1\rightarrow t_r}^{L}(\epsilon^L;c)
\right);c
\right).
\label{eq:pr_traj}
\end{equation}
where $\mathcal{T}_{L \rightarrow H}$ denotes the
cross-resolution transition operator to convert the low-resolution state into a high-resolution state.

\subsection{Analysis of Resolution Transition}
\label{sec:vae_inconsistency}

At transition time $t_r$, we analyze the predicted clean endpoint to expose
decoder-visible differences without the noise retained in the intermediate
state:
\begin{equation}
\widehat z_{0,t_r}^{\mathrm L}
=
z_{t_r}^{\mathrm L}
-
t_r v_\theta
\left(
z_{t_r}^{\mathrm L},t_r,c
\right).
\label{eq:clean_endpoint}
\end{equation}
Starting from the same estimate, direct latent lifting and pixel-VAE
re-encoding produce

\begin{equation}
z^{\mathrm{lat}}
=
\mathcal U^{\mathrm{lat}}
\left(
\widehat z_{0,t_r}^{\mathrm L}
\right),
\label{eq:latent_route}
\end{equation}

\begin{equation}
Y
=
\mathcal U^{\mathrm{pix}}
\left(
\mathcal D_{\mathrm L}
\left(
\widehat{z}^{\mathrm L}_{0,t_r}
\right)
\right),
\qquad
z^{\mathrm{pix}}
=
\mathcal E_{\mathrm H}(Y),
\label{eq:pixel_route}
\end{equation}
where $\mathcal U^{\mathrm{lat}}$ and $\mathcal U^{\mathrm{pix}}$ denote latent-space and pixel-space resizing. 
$\mathcal E_{\mathrm H}$ and $\mathcal D_{\mathrm L}$ represent the high-resolution encoder and the low-resolution decoder, respectively.

\paragraph{\textbf{Observation 1.}} 
\emph{Decoder-visible inconsistencies under direct latent interpolation.}

\begin{figure}[t]
    \centering
    \includegraphics[width=\columnwidth]{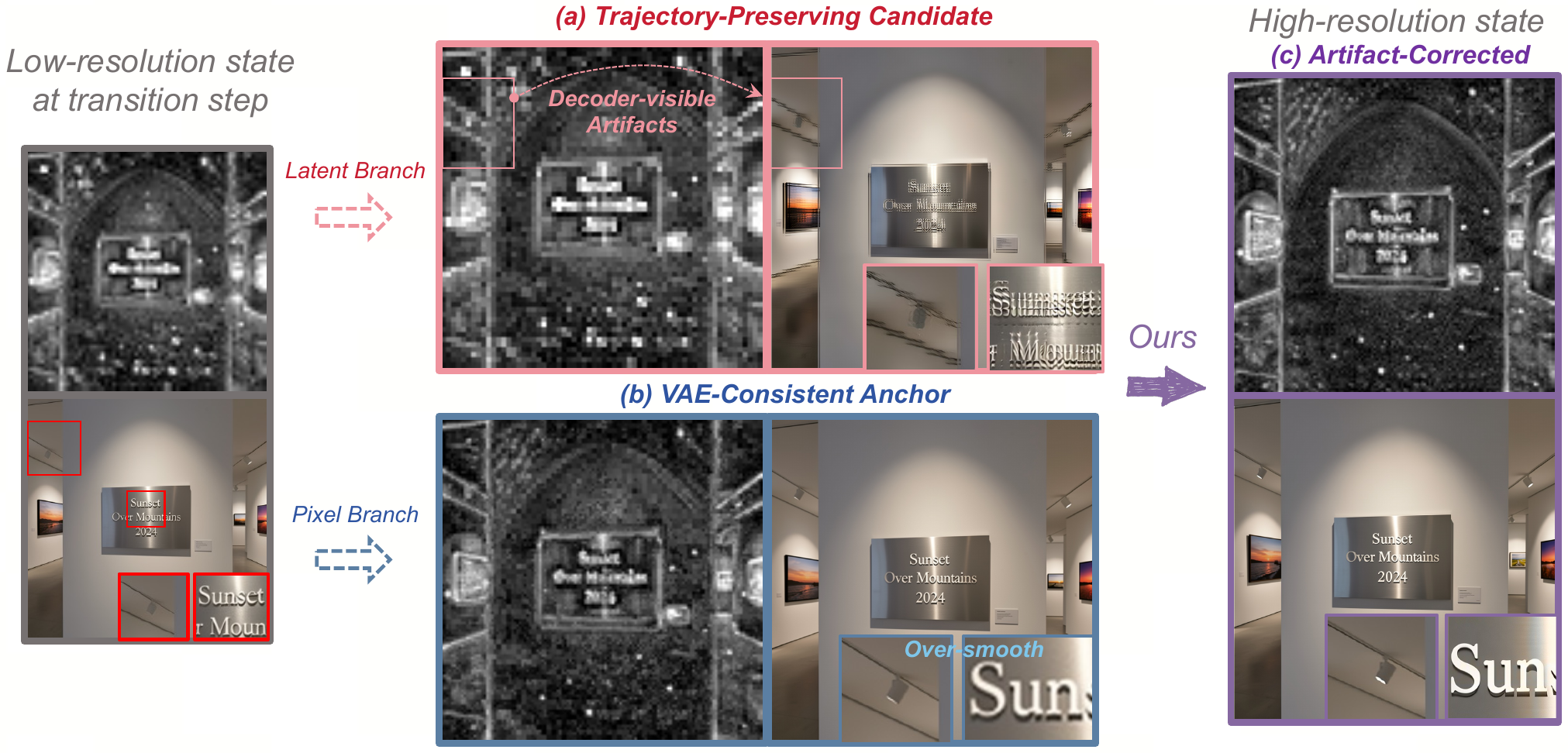}
    \caption{
    Direct lifting introduces decoder-visible inconsistency (a), whereas pixel-VAE
    re-encoding provides a stable yet smoother VAE-reachable anchor (b).
    Correction toward this anchor yields an artifact-reduced
    target-resolution state (c).
}
    \label{fig:vae_consistancy1}
\end{figure}

Let $ \mathcal U^{\mathrm{lat}}:\mathbb R^{d_{\mathrm L}}\rightarrow\mathbb R^{d_{\mathrm H}}$ be a fixed linear interpolation operator, where $d_{\mathrm L}<d_{\mathrm H}$. Its reachable subspace is  $\ReachSubet:=\left\{\mathcal U^{\mathrm{lat}}\left(z^{\mathrm L}\right) \mid z^{\mathrm L}\in\mathbb R^{d_{\mathrm L}}\right\}.$ 
By linearity, $dim(\ReachSubet)=rank(\mathcal U^{\mathrm{lat}})\leq d_{\mathrm L}<d_{\mathrm H}$.
Let $P_{t_r}^{\mathrm{lat}}$ denote the distribution of $z^{\mathrm{lat}}$, and let $P_{\mathrm{VAE}}^{\mathrm H}$ denote the native high-resolution VAE latent distribution. Since every directly lifted latent lies in $\ReachSubet$,
$P_{t_r}^{\mathrm{lat}}(\ReachSubet)=1$. If native
target-resolution VAE latents are not concentrated on this subspace,
i.e., $P_{\mathrm{VAE}}^{\mathrm H}(\ReachSubet)<1$, then
$P_{t_r}^{\mathrm{lat}}\neq P_{\mathrm{VAE}}^{\mathrm H}$.
This support mismatch provides a structural indication of the
distributional discrepancy introduced by direct interpolation. Appendix~B further shows that directly
lifted and pixel latents exhibit $18.93\times$ and
$1.43\times$ the native target-VAE round-trip energy, respectively.
Together with Fig.~\ref{fig:vae_consistancy1}(a), this empirically
associates round-trip instability with ghosting, structural distortion,
and spatial drift. Meanwhile, direct lifting preserves deterministic
correspondence with the preceding low-resolution state, motivating
its use as the trajectory-preserving candidate.

\paragraph{\textbf{Observation 2.}} 
\emph{Pixel re-encoding yields an encoder-reachable representation with limited high-frequency detail.}

Let $\mathcal X_{\mathrm H}$ denote the high-resolution image space and define the encoder-reachable latent set as $\mathcal Z_{\mathrm{pix}}^{\mathrm H}
:=\left\{\mathcal E_{\mathrm H}(x) \mid x\in\mathcal X_{\mathrm H}\right\}$. Although $z^{\mathrm{pix}}\in\mathcal Z_{\mathrm{pix}}^{\mathrm H}$
guarantees encoder reachability, it does not imply that $Y$ uniquely
identifies the corresponding sample-specific high-resolution detail. Let $r_{\mathrm{HF}}$ denote the sample-specific high-frequency component. The minimum MSE attainable by any deterministic predictor
from $Y$ is
\begin{equation}
\inf_f
\mathbb E
\left[
\left\|
r_{\mathrm{HF}}-f(Y)
\right\|_2^2
\right]
=
\mathbb E
\left[
\operatorname{tr}
\operatorname{Cov}
\left(
r_{\mathrm{HF}}\mid Y
\right)
\right].
\label{eq:mse}
\end{equation}

When the conditional distribution $p(r_{\mathrm{HF}}\mid Y)$ is non-degenerate, Eq.~\ref{eq:mse} implies a strictly positive irreducible error. Hence, the pixel–VAE re-encoding branch cannot uniquely recover the sample-specific high-resolution detail from $Y$. 
Figure ~\ref{fig:vae_consistancy1}(b) further shows that the reconstruction of pixel-VAE re-encoding retains coarse spatial information after resolution transitions but produces a smoother result, indicating that it is more suitable for selective correction than as a global alternative to $z^{\mathrm{lat}}$.

\begin{figure*}[!t]
    \centering
    \includegraphics[width=\textwidth]{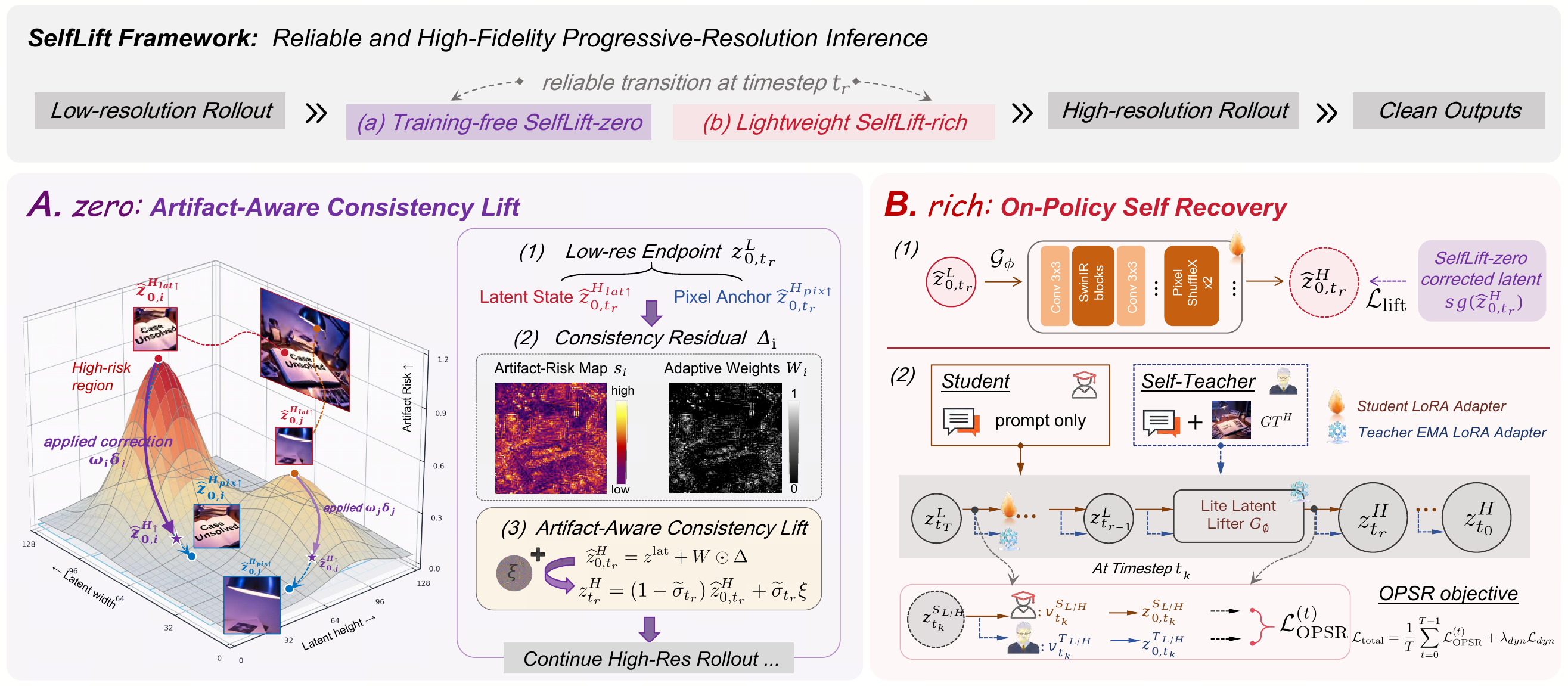}
    \caption{
    \textbf{Overview of SelfLift.}
    Given the predicted low-resolution clean endpoint, \emph{SelfLift-zero}
    constructs a trajectory-preserving latent candidate and a VAE-reachable pixel anchor. Their consistency residual yields an artifact-risk map and adaptive weights for selectively correcting high-risk regions before re-noising.
    \emph{SelfLift-rich} distills this correction into a lightweight latent lifter and performs on-policy recovery on student-visited states, using an EMA self-teacher for privileged high-resolution guidance and a dynamics
    objective to preserve the pretrained few-step trajectory.
}
    \label{fig:overview}
\end{figure*}

\subsection{SelfLift}

To address the two transition failure modes identified above,
we introduce SelfLift, which has two variants. SelfLift-zero uses Artifact-Aware Consistency Lift, which can selectively combine the dual lifting branches without training. SelfLift-rich enables On-Policy Self Recovery to learn a lightweight lifter on the student's progressive-resolution rollout. Figure~\ref{fig:overview} illustrates an overview of our SelfLift framework.

\paragraph{\textbf{Artifact-Aware Consistency Lift}}
\label{sec:artifact_aware_lift}

SelfLift-zero turns the complementary transition behaviors identified in
Sec.~\ref{sec:vae_inconsistency} into an asymmetric, training-free correction operator. 
Treating the two routes symmetrically is suboptimal: replacing the direct-lift state would inherit the smoothing of the pixel route, while global mixing would modify reliable and corrupted regions indiscriminately. 
We therefore retain $z^{\mathrm{lat}}$ as the trajectory state and use $z^{\mathrm{pix}}$ only as a paired correction anchor.
Both routes originate from the same predicted clean endpoint
$\widehat z_{0,t_r}^{L}$, allowing their disagreement to isolate the effect of the resolution transition. 

We define the consistency residual as
$\Delta=z^{\mathrm{pix}}-z^{\mathrm{lat}}$. $\Delta_i$ points from the direct-lift state toward its paired VAE-reachable anchor, while its magnitude reflects their local discrepancy. We convert this discrepancy into a spatial artifact-risk map and select the top-$\rho$ locations within each sample:
\begin{equation}
s_i
=
\frac{1}{C}\lVert\Delta_i\rVert_1,
\qquad
M_i
=
\mathbf{1}\!\left[
s_i
\geq
Q_{1-\rho}\!\left(\{s_j\}_j\right)
\right],
\label{eq:artifact_risk}
\end{equation}
where $Q_{1-\rho}$ denotes the $(1-\rho)$-quantile. Here, $s_i$ serves as a local inconsistency proxy, while the sample-wise quantile adapts the selected region to different prompts and transition steps. We further convert the
selected scores into spatially adaptive correction weights:
\begin{equation}
W_i
=
M_i
\left[
w_{\min}
+
\left(w_{\max}-w_{\min}\right)
\frac{s_i-s_{\min}}
{s_{\max}-s_{\min}+\varepsilon}
\right],
\label{eq:artifact_weight}
\end{equation}
where $s_{\min}$ and $s_{\max}$ are computed over the selected region. The corrected target-resolution clean latent is obtained through \emph{Artifact-Aware Consistency Lift}:
\begin{equation}
\widehat z_{0,t_r}^{H}
=
z^{\mathrm{lat}}
+
W\odot\Delta
=
(1-W)\odot 
z
^{\mathrm{lat}}
+
W\odot z^{\mathrm{pix}}.
\label{eq:consistency_lift}
\end{equation}
Reliable locations remain on the direct-lift trajectory, whereas locations with larger inconsistency receive stronger correction toward the pixel-VAE anchor. 
This selective correction reduces artifacts such as ghosting and spatial drift without globally inheriting the over-smoothing of pixel-VAE re-encoding.
The corrected result is shown in Figure~\ref{fig:vae_consistancy1}(c).

Finally, we re-noise the corrected clean estimate at $t_r$ under the original flow marginal:
\begin{equation}
z_{t_r}^{H}
=
\left(1-\widetilde{\sigma}_{t_r}\right)
\widehat z_{0,t_r}^{H}
+
\widetilde{\sigma}_{t_r}\xi,
\qquad
\xi\sim\mathcal N(0,I).
\label{eq:transition_renoise}
\end{equation}

Sampling then resumes at the target resolution under the unchanged few-step schedule. \emph{SelfLift-zero} is training-free and plug-and-play, requiring neither external super-resolution nor additional denoiser evaluations beyond the original sampling schedule.

\begin{figure*}[!t]
    \centering
    \includegraphics[
        width=0.96\textwidth
    ]{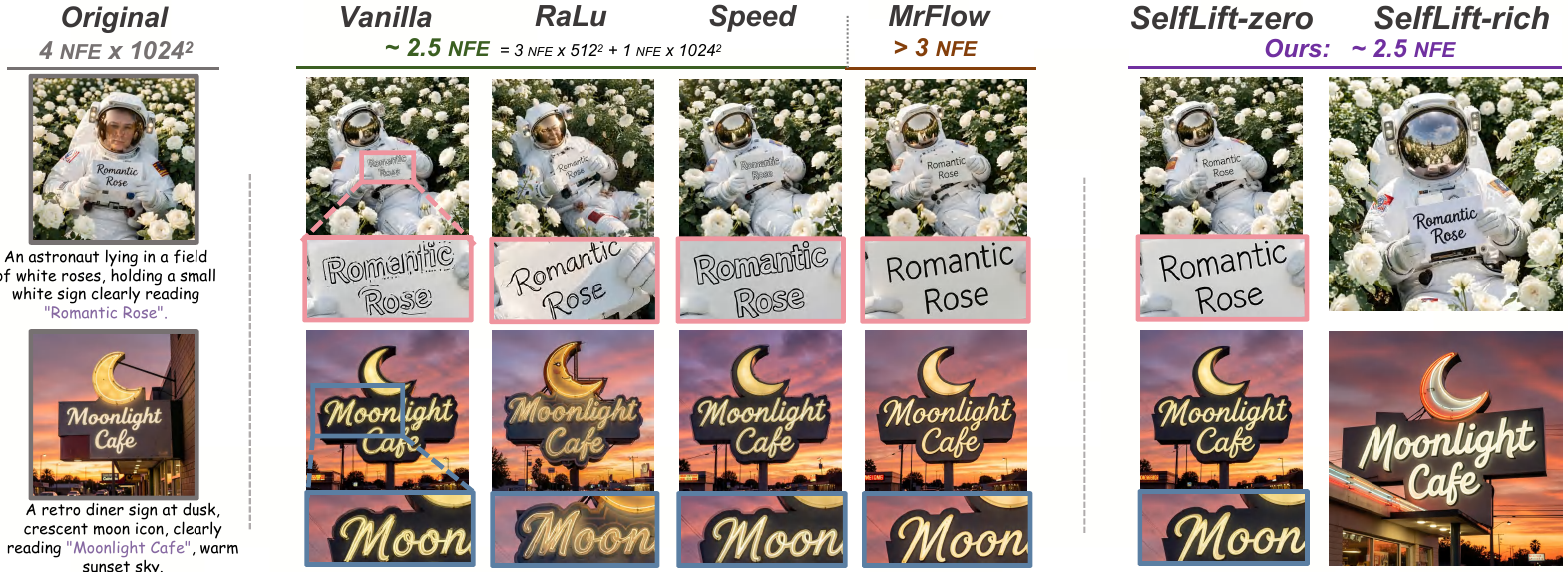}
    \caption{
    \textbf{Superior visual fidelity of \emph{SelfLift} under aggressive acceleration on FLUX.2-Klein-9b (4 NFEs).}
    Vanilla, RALU, and Speed show severe artifacts, while MrFlow improves fidelity using a costlier external super-resolution model.
    \textbf{\emph{SelfLift-zero}} suppresses artifacts through model-native correction, and \textbf{\emph{SelfLift-rich}} further restores richer details.
}
    \vspace{-0.3cm}
    \label{fig:qualitative_comparison}
\end{figure*}

\paragraph{\textbf{On-Policy Self Recovery}}
\label{sec:on_policy_self_recovery}

SelfLift-zero provides stable resolution transition and substantial
training-free acceleration. SelfLift-rich further advances the
efficiency--quality frontier.
Although one VAE round trip is inexpensive, its fixed cost becomes more
noticeable when sampling requires only a few denoiser evaluations. 
We therefore
distill the complete Artifact-Aware Consistency Lift into a compact latent
lifter $G_{\phi}$. For a randomly sampled valid transition step $t_r$, we parameterize the lifter as a residual update over direct latent lifting:
\begin{equation}
\mathcal{G}_{\phi}\!\left(\hat{z}^{L}_{0,t_r}\right)
=
z^{\mathrm{lat}} + \mathcal{R}_{\phi}\!\left(\hat{z}^{L}_{0,t_r}\right),
\end{equation}
\begin{equation}
\mathcal{L}_{\mathrm{lift}}
=
\mathbb{E}_{c,t_r}
\left[
\left\|
\mathcal{G}_{\phi}\!\left(\hat{z}^{L}_{0,t_r}\right)
-
\mathrm{sg}\!\left(\hat{z}^{H}_{0,t_r}\right)
\right\|_2^2
\right],
\end{equation}
where $\mathrm{sg}(\cdot)$ denotes stop-gradient. 
The residual parameterization preserves the state established by the low-resolution prefix and learns only the required transition correction. We instantiate $\mathcal{R}_{\phi}$ with a compact SwinIR-style network. The resulting 10.8M-parameter lifter replaces the explicit decode--resize--encode path with a single latent-space forward pass, yielding
further end-to-end acceleration.

We then freeze $\mathcal G_{\phi}$ and train directly on the
progressive-resolution trajectory generated by the student. At each
student-visited state $z_{t_k}^{\mathrm S}$, the student with parameters $\theta$
predicts $\widehat z_{0,t_k}^{\mathrm S}$ from the text condition $c$, while an
EMA self-teacher with parameters $\bar{\theta}$ predicts
$\widehat z_{0,t_k}^{\mathrm T}$ using additional high-resolution privileged
context. Both predictions are evaluated on the same state, avoiding supervision
from an independent full-resolution trajectory.
Let $\Phi_{\theta,t_k}(z;c)$ denote the one-step sampling update from $t_k$ to
$t_{k+1}$. We optimize the student using
\begin{equation}
\label{eq:opsr_objective}
\begin{aligned}
\mathcal L_{\mathrm{rich}}
&=
\mathcal L_{\mathrm{OPSR}}
+\lambda_{\mathrm{dyn}}\mathcal L_{\mathrm{dyn}},
\\[0.35em]
\mathcal L_{\mathrm{OPSR}}
&=
\frac{1}{K}\sum_{k=1}^{K}
\left\|
\widehat z_{0,t_k}^{\mathrm S}
-
\operatorname{sg}\!\left(
\widehat z_{0,t_k}^{\mathrm T}
\right)
\right\|_2^2,
\\[0.35em]
\mathcal L_{\mathrm{dyn}}
&=
\frac{1}{K}\sum_{k=1}^{K}
\left\|
\begin{aligned}
&\Phi_{\theta,t_k}
 \!\left(z_{t_k}^{\mathrm S};c\right)
\\[-0.15em]
&\quad-
 \operatorname{sg}\!\left(
 \Phi_{\theta_0,t_k}
 \!\left(z_{t_k}^{\mathrm S};c\right)
 \right)
\end{aligned}
\right\|_2^2 .
\end{aligned}
\end{equation}

where $K$ is the number of supervised denoising steps and $\theta_0$ denotes
the frozen pretrained backbone. $\mathcal L_{\mathrm{OPSR}}$ transfers dense
high-resolution guidance on the student-visited states, while
$\mathcal L_{\mathrm{dyn}}$ preserves the pretrained few-step dynamics. The
rollout is advanced only by the student, with gradients stopped across
denoising steps.

At inference, the privileged high-resolution context and EMA self-teacher are
removed. \emph{SelfLift-rich} retains only the adapted
prompt-conditioned model and the lightweight latent lifter.
Complete inference and training procedures are provided in Appendix~A.

\providecommand{\slmetricdelta}[3]{%
  \makebox[3.2em][r]{#1}%
  \,\makebox[3.0em][l]{{\scriptsize\textcolor{#2}{(#3)}}}%
}
\providecommand{\slmetricref}[1]{%
  \makebox[3.2em][r]{#1}%
  \,\makebox[3.0em][l]{}%
}
\providecommand{\slspeeddelta}[3]{%
  \makebox[4.3em][r]{#1}%
  \,\makebox[3.8em][l]{{\scriptsize\textcolor{#2}{(#3)}}}%
}
\providecommand{\slspeedref}[1]{%
  \makebox[4.3em][r]{#1}%
  \,\makebox[3.8em][l]{}%
}

\begin{table*}[!t]
\centering

\caption{
\textbf{End-to-end efficiency and generation quality on FLUX.2-Klein-9b and Z-Image-Turbo.}
Speedups ($\times$) are relative to Base (50), while parenthesized percentages denote latency reductions from Base (4)/(8).
Other parenthesized values are changes from SelfLift-zero.
\textcolor{green!50!black}{Green} marks SelfLift improvements, and \textcolor{red}{red} marks the two lowest latency reductions or two largest quality drops among competing methods.
Bold and underline denote the best and second-best non-Base quality results.
$\uparrow$/$\downarrow$ indicate higher/lower is better.
}
\label{tab:main_results}

\resizebox{\textwidth}{!}{
\begin{tabular}{l|cc|c|cc|cc}

\toprule

\multicolumn{1}{c|}{\multirow{2}{*}{Method}}
&
\multicolumn{2}{c|}{Acceleration}
&
\multicolumn{1}{c|}{Overall}
&
\multicolumn{2}{c|}{Image Quality}
&
\multicolumn{2}{c}{Text Alignment}
\\

\cmidrule(lr){2-3}
\cmidrule(lr){4-4}
\cmidrule(lr){5-6}
\cmidrule(lr){7-8}

&
Latency (s)$\downarrow$
&
Speedup$\uparrow$
&
ImageReward$\uparrow$
&
PickScore$\uparrow$
&
AestheticScore$\uparrow$
&
CLIP$\uparrow$
&
GenEval$\uparrow$
\\

\midrule
\multicolumn{8}{c}{\textbf{(a) FLUX.2-Klein-9b}~\cite{blackforestlabs2026flux2klein}}
\\
\midrule

Base (50)
&46.053&\slspeedref{1.000$\times$}
&
\slmetricref{1.129}
&
\slmetricref{22.330}
&
\slmetricref{5.465}
&
\slmetricref{32.039}
&
\slmetricref{0.784}

\\

Base (4)
&2.665&\slspeeddelta{17.280$\times$}{gray}{+0.0\%}
&
\slmetricref{1.312}
&
\slmetricref{22.729}
&
\slmetricref{5.647}
&
\slmetricref{32.342}
&
\slmetricref{0.879}

\\

\midrule

Vanilla Cache (2) ~\cite{liu2025teacache}
&1.520&\slspeeddelta{30.310$\times$}{gray}{+43.0\%}
&
\slmetricdelta{1.176}{red}{-0.125}
&
\slmetricdelta{22.254}{red}{-0.379}
&
\slmetricdelta{5.586}{red}{-0.062}
&
\slmetricdelta{32.164}{gray}{-0.153}
&
\slmetricdelta{0.805}{red}{-0.048}

\\

Vanilla Dyres
&1.512&\slspeeddelta{30.470$\times$}{gray}{+43.3\%}
&
\slmetricdelta{1.278}{gray}{-0.023}
&
\slmetricdelta{22.504}{gray}{-0.129}
&
\slmetricdelta{5.601}{red}{-0.047}
&
\slmetricdelta{31.799}{red}{-0.518}
&
\slmetricdelta{0.782}{red}{-0.071}

\\

LSRNA-RNA~\cite{jeong2025lsrna}
&1.660&\slspeeddelta{27.740$\times$}{gray}{+37.7\%}
&
\slmetricdelta{1.254}{red}{-0.047}
&
\slmetricdelta{22.494}{red}{-0.139}
&
\slmetricdelta{5.606}{gray}{-0.042}
&
\slmetricdelta{31.749}{red}{-0.568}
&
\slmetricdelta{0.868}{gray}{+0.015}

\\

Bottleneck~\cite{tian2025bottleneck}
&1.857&\slspeeddelta{24.800$\times$}{red}{+30.3\%}
&
\slmetricdelta{1.277}{gray}{-0.024}
&
\slmetricdelta{22.629}{gray}{-0.004}
&
\slmetricdelta{\underline{5.655}}{gray}{+0.007}
&
\slmetricdelta{32.165}{gray}{-0.152}
&
\slmetricdelta{\textbf{0.901}}{gray}{+0.048}

\\

RaLu~\cite{jeong2025ralu}
&1.662&\slspeeddelta{27.700$\times$}{gray}{+37.6\%}
&
\slmetricdelta{1.289}{gray}{-0.012}
&
\slmetricdelta{22.609}{gray}{-0.024}
&
\slmetricdelta{5.621}{gray}{-0.027}
&
\slmetricdelta{32.176}{gray}{-0.141}
&
\slmetricdelta{0.847}{gray}{-0.006}

\\

Speed~\cite{xiao2026spectral}
&1.601&\slspeeddelta{28.770$\times$}{gray}{+39.9\%}
&
\slmetricdelta{1.292}{gray}{-0.009}
&
\slmetricdelta{22.583}{gray}{-0.050}
&
\slmetricdelta{5.638}{gray}{-0.010}
&
\slmetricdelta{32.243}{gray}{-0.074}
&
\slmetricdelta{0.850}{gray}{-0.003}

\\

MrFlow~\cite{zheng2026mrflow}
&2.165&\slspeeddelta{21.270$\times$}{red}{+18.8\%}
&
\slmetricdelta{1.283}{gray}{-0.018}
&
\slmetricdelta{22.513}{gray}{-0.120}
&
\slmetricdelta{5.601}{gray}{-0.047}
&
\slmetricdelta{32.301}{gray}{-0.016}
&
\slmetricdelta{0.852}{gray}{-0.001}

\\

\midrule

SelfLift-zero
&1.643
&\slspeeddelta{28.040$\times$}{green!50!black}{+38.3\%}
&
\slmetricref{\underline{1.301}}
&
\slmetricref{\underline{22.633}}
&
\slmetricref{5.648}
&
\slmetricref{\underline{32.317}}
&
\slmetricref{0.853}

\\

\textbf{SelfLift-rich}
&\textbf{1.556}
&\slspeeddelta{\textbf{29.610$\times$}}{green!50!black}{+41.5\%}
&
\slmetricdelta{\textbf{1.317}}{green!50!black}{+0.016}
&
\slmetricdelta{\textbf{22.704}}{green!50!black}{+0.071}
&
\slmetricdelta{\textbf{5.688}}{green!50!black}{+0.040}
&
\slmetricdelta{\textbf{32.345}}{green!50!black}{+0.029}
&
\slmetricdelta{\underline{0.886}}{green!50!black}{+0.033}

\\

\midrule

\multicolumn{8}{c}{\textbf{(b) Z-Image-Turbo}~\cite{zimage2025}}
\\
\midrule

Base (50)
&38.562&\slspeedref{1.000$\times$}
&
\slmetricref{1.029}
&
\slmetricref{22.158}
&
\slmetricref{5.533}
&
\slmetricref{32.037}
&
\slmetricref{0.713}

\\

Base (8)
&3.591&\slspeeddelta{10.740$\times$}{gray}{+0.0\%}
&
\slmetricref{1.090}
&
\slmetricref{22.510}
&
\slmetricref{5.536}
&
\slmetricref{32.091}
&
\slmetricref{0.754}

\\

\midrule

TaylorSeer~\cite{liu2025taylorseer}
&2.432&\slspeeddelta{15.850$\times$}{gray}{+32.3\%}
&
\slmetricdelta{\underline{1.081}}{gray}{+0.046}
&
\slmetricdelta{22.430}{gray}{-0.004}
&
\slmetricdelta{5.463}{gray}{-0.012}
&
\slmetricdelta{31.880}{gray}{-0.096}
&
\slmetricdelta{0.728}{gray}{-0.021}

\\

Vanilla Dyres
&2.585&\slspeeddelta{14.920$\times$}{red}{+28.0\%}
&
\slmetricdelta{1.011}{gray}{-0.024}
&
\slmetricdelta{22.336}{gray}{-0.098}
&
\slmetricdelta{5.373}{red}{-0.102}
&
\slmetricdelta{31.302}{red}{-0.674}
&
\slmetricdelta{0.716}{red}{-0.033}

\\

LSRNA-RNA~\cite{jeong2025lsrna}
&2.382&\slspeeddelta{16.190$\times$}{gray}{+33.7\%}
&
\slmetricdelta{0.993}{red}{-0.042}
&
\slmetricdelta{22.050}{red}{-0.384}
&
\slmetricdelta{5.389}{red}{-0.086}
&
\slmetricdelta{31.015}{red}{-0.962}
&
\slmetricdelta{0.682}{red}{-0.067}

\\

Bottleneck~\cite{tian2025bottleneck}
&2.485&\slspeeddelta{15.520$\times$}{red}{+30.8\%}
&
\slmetricdelta{0.941}{red}{-0.094}
&
\slmetricdelta{22.196}{red}{-0.238}
&
\slmetricdelta{5.439}{gray}{-0.036}
&
\slmetricdelta{31.714}{gray}{-0.263}
&
\slmetricdelta{\underline{0.763}}{gray}{+0.014}

\\

Speed~\cite{xiao2026spectral}
&1.987&\slspeeddelta{19.410$\times$}{gray}{+44.7\%}
&
\slmetricdelta{1.004}{gray}{-0.031}
&
\slmetricdelta{22.282}{gray}{-0.152}
&
\slmetricdelta{5.405}{gray}{-0.070}
&
\slmetricdelta{31.971}{gray}{-0.005}
&
\slmetricdelta{0.743}{gray}{-0.006}

\\

MrFlow~\cite{zheng2026mrflow}
&2.315&\slspeeddelta{16.660$\times$}{gray}{+35.5\%}
&
\slmetricdelta{1.032}{gray}{-0.003}
&
\slmetricdelta{22.380}{gray}{-0.054}
&
\slmetricdelta{\underline{5.501}}{gray}{+0.026}
&
\slmetricdelta{31.945}{gray}{-0.032}
&
\slmetricdelta{0.740}{gray}{-0.009}

\\

\midrule

SelfLift-zero
&2.376
&\slspeeddelta{16.230$\times$}{green!50!black}{+33.8\%}
&
\slmetricref{1.035}
&
\slmetricref{\underline{22.434}}
&
\slmetricref{5.475}
&
\slmetricref{\underline{31.976}}
&
\slmetricref{0.749}

\\

\textbf{SelfLift-rich}
&\textbf{2.007}
&\slspeeddelta{\textbf{19.210$\times$}}{green!50!black}{+44.1\%}
&
\slmetricdelta{\textbf{1.093}}{green!50!black}{+0.058}
&
\slmetricdelta{\textbf{22.499}}{green!50!black}{+0.065}
&
\slmetricdelta{\textbf{5.570}}{green!50!black}{+0.095}
&
\slmetricdelta{\textbf{32.084}}{green!50!black}{+0.107}
&
\slmetricdelta{\textbf{0.770}}{green!50!black}{+0.021}

\\

\bottomrule

\end{tabular}
}

\end{table*}

\begin{table*}[!t]
\centering

\caption{
\textbf{Complementarity with temporal acceleration on 50-step backbones.}
$\mathbf{S}$ and $\mathbf{T}$ denote spatial and temporal acceleration, respectively.
SelfLift compounds acceleration to $5.60\times$/$4.58\times$ while keeping every quality drop below $5\%$ versus Base (50).
}
\label{tab:temporal_composition}

{
\setlength{\tabcolsep}{2.8pt}
\renewcommand{\arraystretch}{1.00}

\resizebox{0.96\textwidth}{!}{
\begin{tabular}{>{\centering\arraybackslash}p{10em}|l|c|cc|c|cc|cc}

\toprule

\multicolumn{1}{c|}{\multirow{2}{*}{Model}}
&
\multicolumn{1}{c|}{\multirow{2}{*}{Method}}
&
\multicolumn{1}{c|}{\multirow{2}{*}{Accel.}}
&
\multicolumn{2}{c|}{Acceleration}
&
\multicolumn{1}{c|}{Overall}
&
\multicolumn{2}{c|}{Image Quality}
&
\multicolumn{2}{c}{Text Alignment}
\\

\cmidrule(lr){4-5}
\cmidrule(lr){6-6}
\cmidrule(lr){7-8}
\cmidrule(lr){9-10}

&
&
&
Latency (s)$\downarrow$
&
Speedup$\uparrow$
&
ImageReward$\uparrow$
&
PickScore$\uparrow$
&
AestheticScore$\uparrow$
&
CLIP$\uparrow$
&
GenEval$\uparrow$
\\

\midrule

\rowcolor{red!5}
\cellcolor{white}
& SelfLift-zero
& $\mathbf{S}$
& 23.966
& 1.920$\times$
& 1.110
& 22.344
& 5.456
& 32.297
& 0.817
\\

\multirow{3}{=}{\centering\raisebox{0.5\baselineskip}{\makecell{\textbf{FLUX.2}\\Klein-9b}}}
& \quad + TeaCache
& $\mathbf{S+T}$
& 11.093
& 4.150$\times$
& 1.119
& 22.356
& 5.474
& 31.994
& 0.774
\\

& \quad + TaylorSeer ($N=3$)
& $\mathbf{S+T}$
& 11.130
& 4.140$\times$
& \textbf{1.120}
& 22.329
& 5.452
& 32.120
& 0.812
\\

& \quad + TaylorSeer ($N=5$)
& $\mathbf{S+T}$
& \textbf{8.225}
& \textbf{5.600$\times$}
& \slmetricdelta{1.102}{green!50!black}{$-2.39\%$}
& \slmetricdelta{22.249}{green!50!black}{$-0.36\%$}
& \slmetricdelta{5.421}{green!50!black}{$-0.81\%$}
& \slmetricdelta{31.805}{green!50!black}{$-0.73\%$}
& \slmetricdelta{0.782}{green!50!black}{$-0.26\%$}
\\

\midrule

\rowcolor{red!5}
\cellcolor{white}
& SelfLift-zero
& $\mathbf{S}$
& 17.689
& 2.180$\times$
& 1.013
& 22.044
& 5.540
& 32.083
& 0.717
\\

\multirow{3}{=}{\centering\raisebox{0.5\baselineskip}{\makecell{\textbf{Z-Image}}}}
& \quad + TeaCache
& $\mathbf{S+T}$
& 10.107
& 3.820$\times$
& 1.022
& 22.136
& 5.564
& 31.960
& 0.695
\\

& \quad + TaylorSeer ($N=3$)
& $\mathbf{S+T}$
& 10.752
& 3.590$\times$
& \textbf{1.032}
& 22.166
& 5.536
& 32.020
& 0.701
\\

& \quad + TaylorSeer ($N=5$)
& $\mathbf{S+T}$
& \textbf{8.424}
& \textbf{4.580$\times$}
& \slmetricdelta{1.018}{green!50!black}{$-1.07\%$}
& \slmetricdelta{22.078}{green!50!black}{$-0.36\%$}
& \slmetricdelta{5.493}{green!50!black}{$-0.72\%$}
& \slmetricdelta{31.810}{green!50!black}{$-0.71\%$}
& \slmetricdelta{0.682}{green!50!black}{$-4.35\%$}
\\

\bottomrule

\end{tabular}
}
}

\end{table*}

\section{Experiments}
\subsection{Settings}
\paragraph{\textbf{Backbones and Inference Protocol.}}
We evaluate SelfLift on FLUX.2-Klein-9b with 4 NFEs and Z-Image-Turbo with 8 NFEs.
All images are generated at $1024\times1024$, with the low-resolution prefix executed at $512\times512$ before a single transition to the target resolution.
We set $t_r=3$ and $t_r=6$, respectively, and use $t_r=35$ for the complementary 50-NFE experiments unless otherwise specified.

\paragraph{\textbf{SelfLift Configurations.}}
For \emph{SelfLift-zero}, we use
$(\rho,w_{\min},w_{\max})=(0.4,0.5,1.0)$ on FLUX.2-Klein-9b and
$(0.3,0.5,1.0)$ on Z-Image-Turbo.
\emph{SelfLift-rich} employs a 10.8M-parameter SwinIR-style latent lifter~\cite{liang2021swinir} trained for 5K iterations with a learning rate of $1\times10^{-4}$.
For On-Policy Self Recovery, we follow D-OPSD's training
configuration~\cite{jiang2026dopsd}, set $\lambda_{\mathrm{dyn}}=80$,
and adopt Flow-GRPO's data protocol~\cite{liu2025flowgrpo}.
Specifically, we sample 40K prompts from the Pick-a-Pic training
split~\cite{kirstain2023pickapic}, a GenEval pool deduplicated against
the official test set~\cite{ghosh2023geneval}, and a visual
text-rendering training set, with probabilities $0.7/0.2/0.1$.
All training is conducted on 8 high-performance GPUs with a per-device batch size of 1.
Additional details, including complete hyperparameter configurations and training costs, are provided in Appendix C.
\begin{figure*}[!t]
    \centering
    \includegraphics[width=\textwidth]{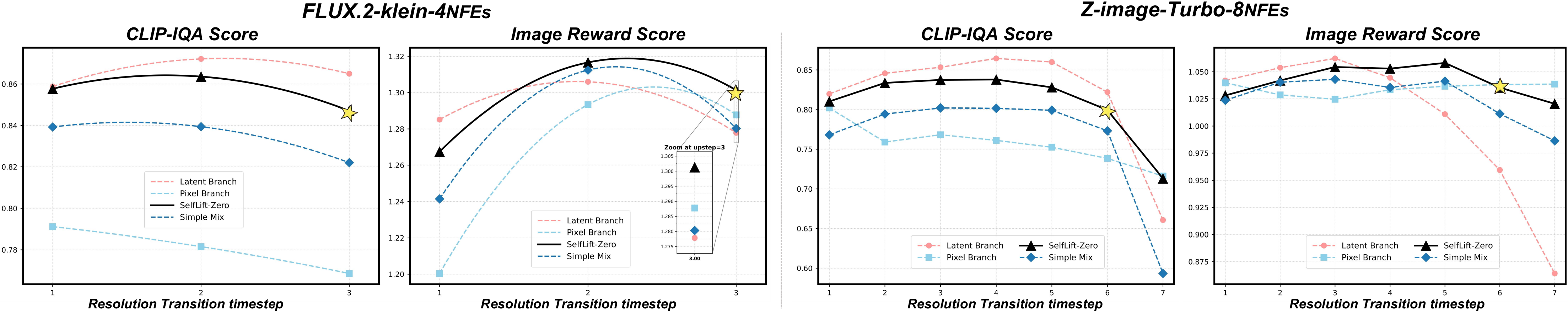}
    \caption{
        Transition-strategy ablation across resolution-transition steps.
        CLIP-IQA~\cite{wang2023clipiqa} measures visual sharpness, while ImageReward evaluates prompt-aligned quality. Stars mark the SelfLift-zero operating points with the best sharpness–alignment trade-off.
    }
    \vspace{-0.2cm}
    \label{fig:ablation1}
\end{figure*}

\paragraph{\textbf{Evaluation Metrics.}}
End-to-end latency is measured on a single GPU under the same hardware setting, using 10 warm-up runs followed by the mean of 50 generations.
Speedup is computed relative to the corresponding 50-NFE backbone.
ImageReward~\cite{xu2023imagereward} measures overall generation quality, PickScore~\cite{kirstain2023pickapic} and AestheticScore~\cite{schuhmann2022laion5b} assess preference and aesthetics, while CLIPScore~\cite{hessel2021clipscore} and GenEval evaluate text--image alignment and compositional instruction following.
Evaluation uses the corresponding test sets.
GenEval uses fixed 300 random official test prompts. All other metrics use an equally sized, predefined balanced set drawn uniformly from DrawBench~\cite{saharia2022imagen}, Pick-a-Pic, and corresponding visual text rendering test split.
All scores are averaged over three shared seeds, yielding 900 paired generations per method.

\begin{table}[t]
\centering
\caption{
Ablation of On-Policy Self Recovery on FLUX.2-Klein-9b.
All learned variants use the same transition step \ensuremath{t_r=3}.
}
\label{tab:rich_ablation}

\setlength{\tabcolsep}{3pt}
\renewcommand{\arraystretch}{1.0}

\resizebox{\columnwidth}{!}{%
\begin{tabular}{l|cc|cc}

\toprule

\multicolumn{1}{c|}{\multirow{2}{*}{Variant}}
&
\multicolumn{2}{c|}{Acceleration}
&
\multicolumn{2}{c}{Quality}
\\

\cmidrule(lr){2-3}
\cmidrule(lr){4-5}

&
\multicolumn{1}{c}{Latency (s)\ensuremath{\downarrow}}
&
\multicolumn{1}{c|}{Speedup\ensuremath{\uparrow}}
&
\multicolumn{1}{c}{ImageReward\ensuremath{\uparrow}}
&
\multicolumn{1}{c}{CLIP\ensuremath{\uparrow}}
\\

\midrule

SelfLift-zero
& 1.642
& 28.04\ensuremath{\times}
& 1.3013
& 32.317
\\

\midrule

- Learned Latent Lifter
& 1.556
& 29.61\ensuremath{\times}
& \underline{1.3031}
& \underline{32.305}
\\

- Off-Policy Distillation
& 1.556
& 29.61\ensuremath{\times}
& 1.1741
& 31.315
\\

- OPSR w/o \ensuremath{\mathcal{L}_{\mathrm{dyn}}}
& 1.556
& 29.61\ensuremath{\times}
& 1.1813
& 31.813
\\

\midrule

\textbf{SelfLift-rich}
& \textbf{1.556}
& \textbf{29.61}\ensuremath{\boldsymbol{\times}}
& \textbf{1.3171}
& \textbf{32.345}
\\

\bottomrule

\end{tabular}
}%
\end{table}

\begin{figure}[t]
    \centering
    \includegraphics[width=0.99\columnwidth]{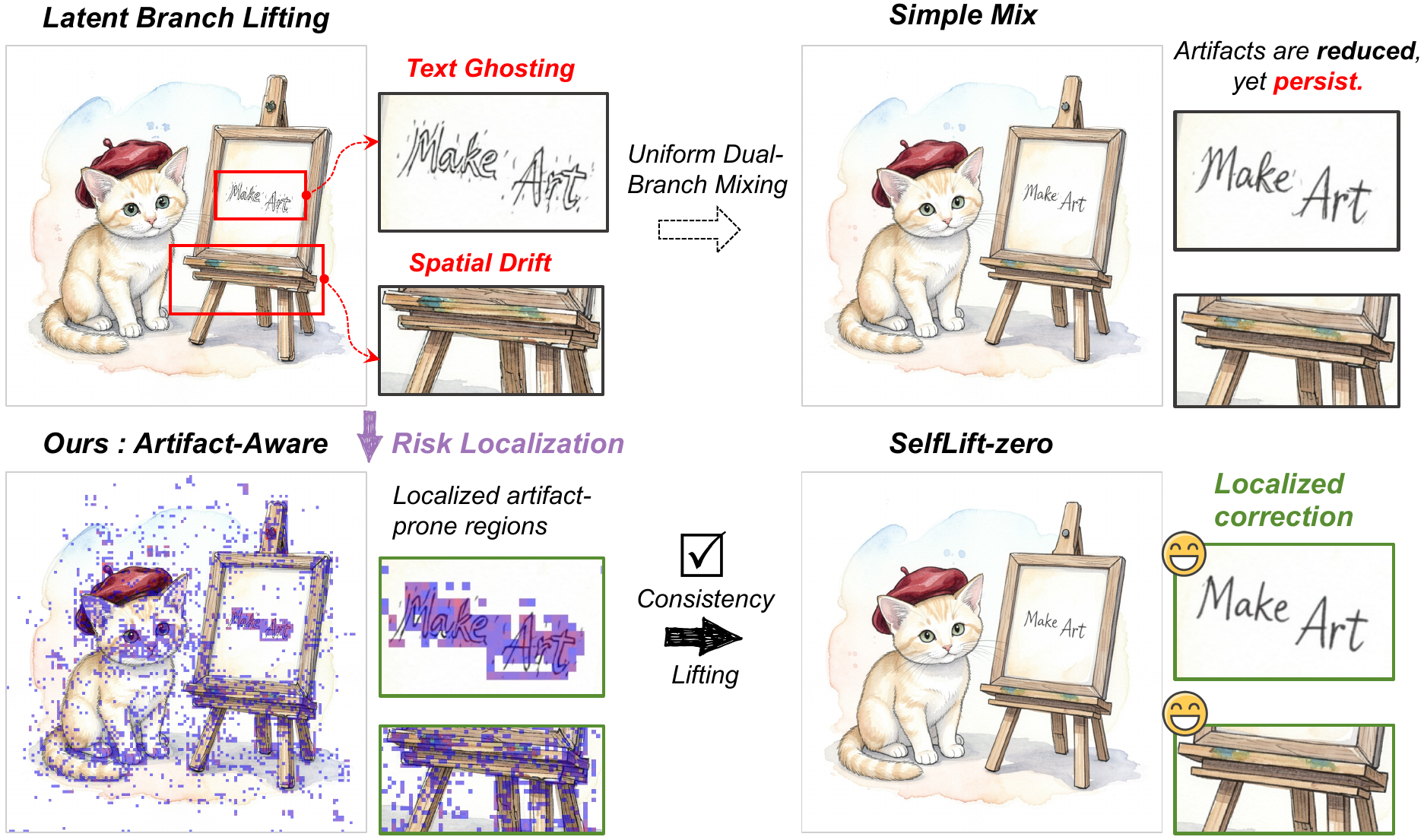}
    \caption{
        Artifact localization and selective correction.
        Uniform mixing only attenuates ghosting and drift, whereas SelfLift-zero corrects localized high-risk regions while preserving reliable details.
    }
    \label{fig:ablation1_2}
\end{figure}

\subsection{Main Results}

\paragraph{\textbf{Few-Step Results.}}
We compare SelfLift with representative temporal acceleration methods, including TeaCache and TaylorSeer, and spatial acceleration methods, including Bottleneck, RaLu, Speed, and MrFlow.
Table~\ref{tab:main_results} and Figure~\ref{fig:qualitative_comparison} summarize the few-step results.
Existing progressive-resolution methods often suffer from transition artifacts and spatial drift, while MrFlow improves fidelity using an external super-resolution model at lower efficiency.
In contrast, training-free \emph{SelfLift-zero} achieves $28.04\times$ and $16.23\times$ speedups on FLUX.2-Klein-9b and Z-Image-Turbo while maintaining strong overall quality.
\emph{SelfLift-rich} further reduces the native few-step latency by $41.5\%$ and $44.1\%$, reaching $29.61\times$ and $19.21\times$ overall speedups.
It consistently improves all reported metrics over \emph{SelfLift-zero} and achieves the best ImageReward, PickScore, AestheticScore, and CLIPScore among accelerated methods on both backbones.
These results show that SelfLift provides a favorable speed-quality trade-off for aggressive few-step inference.

\paragraph{\textbf{Generality and Orthogonality.}}
Although our main focus is accelerating few-step models, SelfLift-zero is a plug-and-play spatial acceleration method orthogonal to temporal acceleration. As a complementary evaluation, Table~\ref{tab:temporal_composition} applies it directly to 50-step backbones and combines it with TeaCache and TaylorSeer. SelfLift-zero alone achieves $1.92\times$/$2.18\times$ speedups, increasing
to $5.60\times$/$4.58\times$ with temporal acceleration, while keeping the
degradation across all evaluation metrics below $5\%$ relative to the
corresponding base models. This strong multiplicative gain shows that SelfLift provides complementary spatial efficiency, enabling additional acceleration across both distilled and standard diffusion models. Further experiments are provided in the Appendix D.

\subsection{Ablation Study}
We conduct controlled ablations of SelfLift's two key principles: spatially adaptive
transition correction in SelfLift-zero and trajectory-aligned self-recovery in
SelfLift-rich. Together, they suppress transition artifacts, recover
high-resolution details, and preserve few-step dynamics.

\paragraph{\textbf{Ablation of Artifact-Aware Consistency Lift.}}
We examine whether reliable resolution transition can be achieved by either transition route alone or by fixed global fusion. Figure~\ref{fig:ablation1} compares direct latent lifting, pixel-VAE re-encoding, uniform $0.5$ mixing, and SelfLift-zero across transition steps on FLUX.2-Klein-9b and Z-Image-Turbo. Later transitions provide greater acceleration but leave less high-resolution recovery. The pixel branch remains stable but consistently over-smooths the output, resulting in low CLIP-IQA. The latent branch preserves sharpness, yet ghosting and spatial drift increasingly degrade ImageReward at late transitions, particularly on Z-Image-Turbo. Uniform mixing only interpolates between these failures because branch reliability varies spatially. In contrast, SelfLift-zero consistently maintains the strongest joint sharpness and prompt-aligned quality across both backbones. Figure~\ref{fig:ablation1_2} further shows that the consistency residual localizes artifact-prone regions, enabling selective correction while preserving reliable latent details.

\paragraph{\textbf{Ablation of On-Policy Self Recovery.}}
Table~\ref{tab:rich_ablation} isolates the latent lifter, trajectory-aligned supervision, and dynamics preservation.
Replacing the pixel-VAE round trip with the learned lifter reduces latency from $1.642$\,s to $1.556$\,s while preserving quality.
Off-policy distillation substantially degrades ImageReward and CLIPScore, confirming the mismatch between off-policy teacher states and student-visited states.
On-policy supervision alleviates this mismatch but remains insufficient without $\mathcal{L}_{\mathrm{dyn}}$.
The full model improves ImageReward from $1.1813$ to $1.3171$ and CLIPScore from $31.813$ to $32.345$ at the same inference cost, demonstrating that trajectory-aligned supervision and dynamics preservation are jointly necessary.

\section{Limitations and Discussion}
While SelfLift demonstrates competitive empirical performance, its self-recovering design builds on capabilities increasingly available in modern multimodal generative models, including native multi-resolution support and image-conditioned editing. SelfLift-zero is training-free and
plug-and-play across compatible model families. Its applicability and
practical adaptation are discussed further in Appendix~E. The effectiveness of SelfLift-rich also depends on the quality of its internal self-teacher. We mitigate this dependence through high-quality prompt--reference curation before training, enabling the internal teacher to provide reliable high-resolution guidance along student-visited trajectories. Importantly, this reliance on the model itself is also what enables SelfLift's self-contained design, without introducing a separately trained teacher or an external reward model. Developing stronger self-teachers and more effective trajectory-aligned supervision remains an important direction for future work.

\section{Conclusion}

We introduced SelfLift for reliable progressive-resolution inference in
few-step diffusion models. SelfLift-zero selectively corrects transition
artifacts without training or external super-resolution, while SelfLift-rich
learns high-resolution recovery directly on student-visited trajectories.
SelfLift reduces the native few-step latency by 41.5\%/44.1\% on
FLUX.2-Klein-9b and Z-Image-Turbo and, together with timestep distillation,
achieves overall speedups of 29.61$\times$/19.21$\times$ over the
corresponding 50-step models.

\bibliographystyle{plainnat}
\bibliography{main}

\clearpage
\appendix
\section{SelfLift Algorithm}
\label{sec:selflift_algorithm}
SelfLift adopts a unified progressive-resolution sampling procedure with two
transition variants. \emph{SelfLift-zero} performs Artifact-Aware
Consistency Lift, whereas \emph{SelfLift-rich} replaces the transition with the distilled latent lifter
$\mathcal G_{\phi}$ and uses the LoRA-adapted student model. The detailed
procedures for SelfLift inference and SelfLift-rich training are provided in
Algorithm~\ref{alg:selflift_inference} and
Algorithm~\ref{alg:selflift_training}, respectively.

\textbf{SelfLift Inference (Algorithm~\ref{alg:selflift_inference})} follows
the low-resolution trajectory to the transition timestep $t_r$, constructs a
reliable target-resolution state using either artifact-aware correction or the
learned latent lifter $\mathcal G_{\phi}$, re-noises the clean estimate, and
completes the remaining trajectory at high resolution.

\textbf{SelfLift-rich Training (Algorithm~\ref{alg:selflift_training})}
distills the complete \emph{zero} transition into $\mathcal G_{\phi}$,
replacing the explicit pixel--VAE path with a compact latent-space operator.
It then performs On-Policy Self Recovery on student-visited states with a
dynamics objective that preserves the pretrained few-step trajectory.

\begin{algorithm}[!b]
\caption{SelfLift Inference}
\label{alg:selflift_inference}
\small
\begin{algorithmic}[1]
\Require Prompt $c$, sampling schedule
         $\mathcal T=\{(t_k,\widetilde\sigma_{t_k})\}_{k=1}^{T}$,
         transition time $t_r$, mode
         $m\in\{\mathrm{zero},\mathrm{rich}\}$
\Require Pretrained model $v_{\theta_0}$,
         VAE $(\mathcal D_{\mathrm L},
         \mathcal E_{\mathrm H},\mathcal D_{\mathrm H})$
\Require $\mathrm{zero}$ parameters
         $\eta=(\rho,w_{\min},w_{\max})$, trained lifter
         $\mathcal G_{\phi}$ and LoRA-adapted model $v_{\theta}$
\Ensure Generated image $x_0$

\State $\widetilde v\gets v_{\theta_0}$ if $m=\mathrm{zero}$,
       otherwise $\widetilde v\gets v_{\theta}$

\AlgPhase{Low-Resolution Rollout}
\State Sample initial noise $z_1^{\mathrm L}\sim\mathcal N(0,I)$
\State Roll out to $t_r$:
       $z_{t_r}^{\mathrm L}\gets
       \Phi^{\mathrm L}_{1\rightarrow t_r}
       (z_1^{\mathrm L},c,\widetilde v)$

\AlgPhase{Resolution Transition}
\State Predict the clean endpoint:
       $\widehat z_{0,t_r}^{\mathrm L}\gets
       z_{t_r}^{\mathrm L}
       -t_r\widetilde v(z_{t_r}^{\mathrm L},t_r,c)$

\If{$m=\mathrm{zero}$}
    \State Construct paired lifts:
           $(z^{\mathrm{lat}},z^{\mathrm{pix}})$
           \EqNote{Eqs.~(4)--(5)}
    \State Compute consistency residual:
           $\Delta\gets z^{\mathrm{pix}}-z^{\mathrm{lat}}$
    \State Compute artifact-risk map $s$, top-$\rho$ mask $M$, and adaptive weights $W$ from $\Delta$ 
    \EqNote{Eqs.~(7)--(8)}
    \State Apply
           $\widehat z_{0,t_r}^{\mathrm H}\gets
           z^{\mathrm{lat}}+W\odot\Delta$
    \EqNote{Eqs.~(9)}
\Else
    \State Apply the distilled lifter:
           $\widehat z_{0,t_r}^{\mathrm H}\gets
           \mathcal G_{\phi}(\widehat z_{0,t_r}^{\mathrm L})$
           
           \EqNote{Eq.~(11)}
\EndIf

\State Sample transition noise $\xi\sim\mathcal N(0,I)$
\State Re-noise at $t_r$:
       $z_{t_r}^{\mathrm H}\gets
       (1-\widetilde\sigma_{t_r})
       \widehat z_{0,t_r}^{\mathrm H}
       +\widetilde\sigma_{t_r}\xi$
       \EqNote{Eq.~(10)}

\AlgPhase{High-Resolution Rollout}
\State Resume sampling:
       $z_0^{\mathrm H}\gets
       \Phi^{\mathrm H}_{t_r\rightarrow0}
       (z_{t_r}^{\mathrm H},c,\widetilde v)$
\State \Return $x_0\gets\mathcal D_{\mathrm H}(z_0^{\mathrm H})$

\end{algorithmic}
\end{algorithm}

\begin{algorithm}[!t]
\caption{SelfLift-rich Training}
\label{alg:selflift_training}
\small
\begin{algorithmic}[1]
\Require Frozen model $v_{\theta_0}$, VAE, prompt set $\mathcal C$
\Require Training set $\mathcal D=\{(c,\mathrm{gt}^{\mathrm H})\}$
\Require Schedule $\mathcal T$, transition set $\mathcal R$,
         zero parameters $\eta$
\Require Dynamics weight $\lambda_{\mathrm{dyn}}$, EMA decay $\beta$
\Ensure Lifter $\mathcal G_{\phi}$ and LoRA-adapted student $v_{\theta}$

\AlgPhase{Stage I, Latent Lifter Distillation}
\State Residual lifter:
       $\mathcal G_{\phi}(\widehat z_{0,t_r}^{\mathrm L})
       \gets z^{\mathrm{lat}}
       +\mathcal R_{\phi}(\widehat z_{0,t_r}^{\mathrm L})$

\For{each lifter-training iteration}
    \State Sample $c\sim\mathcal C$ and $t_r\sim\mathcal R$
    \State Obtain $\widehat z_{0,t_r}^{\mathrm L}$
           from the frozen low-resolution rollout
    \State Construct target
           $\widehat z_{0,t_r}^{\mathrm{H,zero}}$
           with SelfLift-zero
           
           \EqNote{Eqs.~(4)--(9)}
    \State $\mathcal L_{\mathrm{lift}}\gets
           \left\|
           \mathcal G_{\phi}(\widehat z_{0,t_r}^{\mathrm L})
           -
           \operatorname{sg}
           (\widehat z_{0,t_r}^{\mathrm{H,zero}})
           \right\|_2^2$
           \EqNote{Eq.~(12)}
    \State Update $\phi$ using $\nabla_{\phi}\mathcal L_{\mathrm{lift}}$
\EndFor
\State Freeze $\mathcal G_{\phi}$

\AlgPhase{Stage II, On-Policy Self Recovery}
\State Initialize $v_{\theta}\gets v_{\theta_0}$ and
       $v_{\bar\theta}\gets v_{\theta}$

\For{each recovery-training iteration}
    \State Sample $(c,\mathrm{gt}^{\mathrm H})\sim\mathcal D$, $t_r\sim\mathcal R$ and $z_{t_1}^{\mathrm S}\sim\mathcal N(0,I)$

    \For{$k=1,\ldots,K$}
        \If{$t_k=t_r$}
            \State Apply frozen $\mathcal G_{\phi}$ and re-noise
                   the state
        \EndIf

        \State Student endpoint:
               $\widehat z_{0,t_k}^{\mathrm S}\gets
               z_{t_k}^{\mathrm S}
               -t_kv_{\theta}(z_{t_k}^{\mathrm S},t_k,c)$
        \State Teacher endpoint:
        
        \State \hspace{2.0em}%
        $\widehat z_{0,t_k}^{\mathrm T}\gets
               z_{t_k}^{\mathrm S}
               -t_kv_{\bar\theta}
               (z_{t_k}^{\mathrm S},t_k,c,\mathrm{gt}^{\mathrm H})$

        \State 
        $\ell_{\mathrm{OPSR}}^{(k)}\gets
               \left\|
               \widehat z_{0,t_k}^{\mathrm S}
               -
               \operatorname{sg}
               (\widehat z_{0,t_k}^{\mathrm T})
               \right\|_2^2$
        \EqNote{Eq.~(13)}

        \State Student one-step transition:
        
       \State \hspace{2.0em}%
       $z_{t_{k+1}}^{\mathrm S}\gets
       \Phi_{\theta,t_k}(z_{t_k}^{\mathrm S},c)$

        \State Frozen-backbone reference:
        
        \State \hspace{2.0em}%
               $z_{t_{k+1}}^{0}\gets
               \Phi_{\theta_0,t_k}(z_{t_k}^{\mathrm S},c)$
        
        \State $\ell_{\mathrm{dyn}}^{(k)}\gets
               \left\|
               z_{t_{k+1}}^{\mathrm S}
               -
               \operatorname{sg}
               (z_{t_{k+1}}^{0})
               \right\|_2^2$
               \EqNote{Eq.~(13)}
        
        \State Detach before the next timestep:

        \State \hspace{2.0em}%
               $z_{t_{k+1}}^{\mathrm S}\gets
               \operatorname{sg}(z_{t_{k+1}}^{\mathrm S})$
            
        \Comment{stop gradients across steps}
    \EndFor

    \State $\mathcal L_{\mathrm{OPSR}}\gets
           K^{-1}\sum_{k=1}^{K}\ell_{\mathrm{OPSR}}^{(k)}$
    \State $\mathcal L_{\mathrm{dyn}}\gets
           K^{-1}\sum_{k=1}^{K}\ell_{\mathrm{dyn}}^{(k)}$
    \State $\mathcal L_{\mathrm{rich}}\gets
           \mathcal L_{\mathrm{OPSR}}
           +\lambda_{\mathrm{dyn}}\mathcal L_{\mathrm{dyn}}$
           \EqNote{Eq.~(13)}

    \State Update the student LoRA using
           $\nabla_{\theta}\mathcal L_{\mathrm{rich}}$
    \State $\bar\theta\gets
           \beta\bar\theta+(1-\beta)\theta$
\EndFor

\State Discard $v_{\bar\theta}$ and $\mathrm{gt}^{\mathrm H}$
\State \Return $\mathcal G_{\phi}$ and $v_{\theta}$

\end{algorithmic}
\end{algorithm}

\FloatBarrier
\section{Consistency Analysis}
\label{sec:supp_vae_consistency}

Observation~1 in Sec.~3.2 shows that direct latent lifting preserves
continuity with the preceding low-resolution trajectory, but may introduce
decoder-visible inconsistencies due to its structural mismatch with the native
high-resolution VAE latent distribution. The key intuition is that a latent
lying in a stable, VAE-consistent region should be approximately preserved
after a target-VAE decode--encode round trip, whereas an unsupported latent
may undergo an abnormally large change. We examine this phenomenon through
the following calibration experiment.

\paragraph{\textbf{Calibration setup.}}
We construct a small calibration set of $N=1{,}000$ native
$1024\times1024$ images on FLUX.2-Klein. For each image
$x^{\mathrm H}$, we obtain
$x^{\mathrm L}=\mathcal S_{\downarrow}(x^{\mathrm H})$ and
$z^{\mathrm L}=\mathcal E_{\mathrm L}(x^{\mathrm L})$, where
$\mathcal S_{\downarrow}$ downsamples the image to $512\times512$.
We compare the native HR latent
$z^{\mathrm{nat}}=\mathcal E_{\mathrm H}(x^{\mathrm H})$, the directly
lifted latent
$z^{\mathrm{lat}}=\mathcal U^{\mathrm{lat}}(z^{\mathrm L})$, and the
pixel-VAE latent
$z^{\mathrm{pix}}=\mathcal E_{\mathrm H}
(\mathcal U^{\mathrm{pix}}(\mathcal D_{\mathrm L}(z^{\mathrm L})))$.
Here, $z^{\mathrm{nat}}$ serves as the normal round-trip reference,
$z^{\mathrm{lat}}$ uses nearest-neighbor latent lifting, and
$z^{\mathrm{pix}}$ is obtained through low-resolution decoding,
pixel-space resizing, and target-resolution re-encoding.

\paragraph{\textbf{Round-trip metric.}}
For a target-resolution latent
$z\in\mathbb R^{C\times H\times W}$, we define
\begin{equation}
\begin{aligned}
\mathcal P_{\mathrm H}(z)
&=\mathcal E_{\mathrm H}\!\left(\mathcal D_{\mathrm H}(z)\right),\\
e_{\mathrm{VAE}}(z)
&=\frac{1}{CHW}
  \left\|\mathcal P_{\mathrm H}(z)-z\right\|_2^2 .
\end{aligned}
\label{eq:vae_roundtrip_energy}
\end{equation}
Here, $\mathcal P_{\mathrm H}$ performs one target-VAE decode--encode
round trip, and $e_{\mathrm{VAE}}(z)$ measures the resulting latent change.
A small value indicates a stable, VAE-consistent representation, whereas an
abnormally large value indicates a shift toward a VAE-inconsistent region.

\par\medskip
\noindent
\begin{minipage}{\columnwidth}
\centering
\captionof{table}{
Target-VAE round-trip consistency on 1,000 calibration samples.
}
\label{tab:vae_roundtrip_consistency}

\fontsize{8.5pt}{9.5pt}\selectfont
\setlength{\tabcolsep}{3.5pt}
\renewcommand{\arraystretch}{1.08}

\resizebox{\columnwidth}{!}{%
\begin{tabular}{lccc}
\toprule
\textbf{Representation}
& \textbf{Mean $e_{\mathrm{VAE}}$}
& \textbf{/ Native}
& \textbf{Behavior} \\
\midrule
Native HR latent
& 0.01894
& $1.00\times$
& \color{blue}{Stable} \\
Pixel-VAE re-encoding
& 0.02717
& $1.43\times$
& \color{blue}{Stable} \\
Direct latent lifting
& 0.35859
& $18.93\times$
& \color{red}{Abnormal} \\
\bottomrule
\end{tabular}%
}
\end{minipage}
\par\medskip

\begin{figure}[!t]
\centering
\includegraphics[width=\columnwidth]
{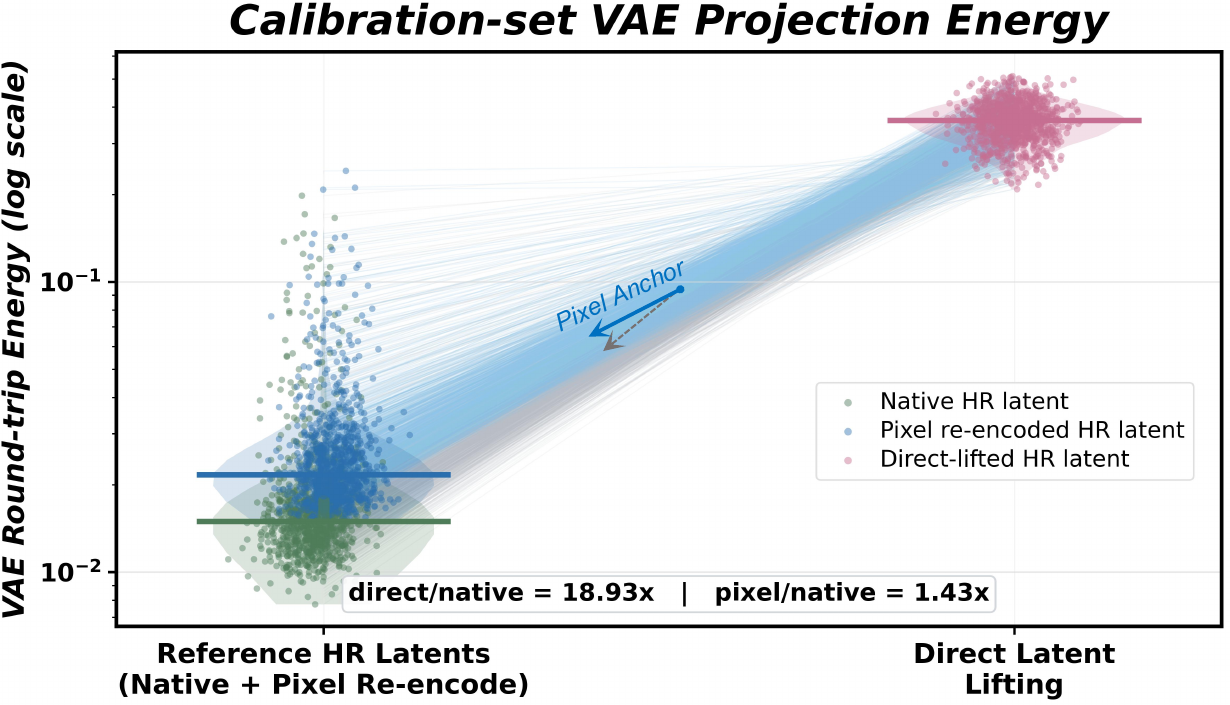}
\caption{
\textbf{Target-VAE round-trip consistency.}
Native HR and pixel-VAE latents remain stable, whereas directly lifted
latents exhibit substantially larger round-trip changes.
}
\label{fig:vae_roundtrip_consistency}
\end{figure}

\paragraph{\textbf{Results.}}
As shown in Table~\ref{tab:vae_roundtrip_consistency} and
Fig.~\ref{fig:vae_roundtrip_consistency}, native HR and pixel-VAE latents
remain stable after the round trip, with small mean energies. In contrast, direct latent lifting yields an abnormally large
energy of $0.35859$, which is $18.93\times$ the native baseline and
$13.20\times$ the pixel-VAE energy. This separation holds for all 1,000
samples, while pixel-VAE re-encoding closes $97.4\%$ of the
direct-lift--native energy gap on average.

These results verify that direct latent lifting shifts the representation
toward a VAE-inconsistent and unstable latent region, rather than merely
changing its spatial resolution. Because the shifted latent is directly
processed by the target-resolution decoder, its unsupported components become
decoder-visible and can manifest as the ghosting, structural distortion, and
spatial drift shown in Fig.~2(a). Pixel-VAE re-encoding instead remains close
to the stable native HR latent regime and provides a practical paired anchor
when the corresponding native HR latent is unavailable during inference.

\section{Experimental Details}
\label{sec:supp_exp_settings}
\begin{table*}[t]
\centering
\caption{
Baseline settings used in the main few-step comparison.
Within the spatial-acceleration block, \cmark\ denotes direct latent lifting,
while \xmark\ is followed by the alternative transition mechanism.
}
\label{tab:baseline_implementation_details}

\fontsize{8.9pt}{9.4pt}\selectfont
\setlength{\tabcolsep}{3.2pt}
\renewcommand{\arraystretch}{1.05}

\begin{adjustbox}{max width=\textwidth}
\begin{tabular}{
@{}
>{\raggedright\arraybackslash}m{1.95cm}
>{\centering\arraybackslash}m{1.55cm}
|>{\raggedright\arraybackslash}m{4.70cm}
|>{\centering\arraybackslash}m{0.65cm}
|>{\raggedright\arraybackslash}m{4.75cm}
|>{\centering\arraybackslash}m{0.65cm}
|>{\raggedright\arraybackslash}m{4.50cm}
@{}
}
\toprule

\multicolumn{1}{c}{
  \multirow{2}{*}{
    {\fontsize{15pt}{13.5pt}\selectfont\textbf{Method}}
  }
}
&
\multicolumn{1}{c}{
  \multirow{2}{*}{
    {\fontsize{15pt}{13.5pt}\selectfont\textbf{Accel.}}
  }
}
&
\multicolumn{1}{|c}{
  \multirow{2}{*}{
    {\fontsize{15pt}{13.5pt}\selectfont\textbf{Core Config.}}
  }
}
&
\multicolumn{2}{|c}{
  {\fontsize{15pt}{13.5pt}\selectfont
   \textbf{FLUX.2-Klein-9b}}
}
&
\multicolumn{2}{|c}{
  {\fontsize{15pt}{13.5pt}\selectfont
   \textbf{Z-Image-Turbo}}
}
\\

\cmidrule(lr){4-5}
\cmidrule(lr){6-7}

&
&
&
\multicolumn{1}{|c}{
  {\fontsize{10pt}{10.5pt}\selectfont\textbf{NFEs}}
}
&
\multicolumn{1}{|c}{
  {\fontsize{10pt}{10.5pt}\selectfont\textbf{Parameters}}
}
&
\multicolumn{1}{|c}{
  {\fontsize{10pt}{10.5pt}\selectfont\textbf{NFEs}}
}
&
\multicolumn{1}{|c}{
  {\fontsize{10pt}{10.5pt}\selectfont\textbf{Parameters}}
}
\\

\midrule

\multicolumn{7}{l}{
\textcolor{blue}{\itshape Reference Models}
}
\\
\addlinespace[1pt]

Base (full)
&
--
&
Native full-resolution inference.
&
50
&
Full $1024^2$ resolution.
&
50
&
Full $1024^2$ resolution.
\\

Base (distilled)
&
--
&
Native few-step inference.
&
4
&
Original 4-NFE schedule.
&
8
&
Original 8-NFE schedule.
\\

\midrule

\multicolumn{7}{l}{
\textcolor{blue}{\itshape Temporal Acceleration}
}
\\
\addlinespace[1pt]

Vanilla Cache
&
\textbf{T}
&
TeaCache-style feature reuse.
&
4
&
Cache steps $1$ and $3$.
&
--
&
--
\\

TaylorSeer
&
\textbf{T}
&
Taylor-based feature forecasting.
&
--
&
--
&
8
&
Official main-table setting.
\\

\midrule

\multicolumn{7}{l}{
\textcolor{blue}{\itshape Spatial Acceleration}
}
\\
\addlinespace[1pt]

Vanilla DyRes
&
\cmark
&
LR prefix with direct latent lifting.
&
4
&
$512^2\!\rightarrow\!1024^2$, $t_r=3$, shift $0.97$.
&
8
&
$512^2\!\rightarrow\!1024^2$, $t_r=6$.
\\

LSRNA-RNA
&
\cmark
&
Canny-guided RNA after latent lifting.
&
4
&
$t_r=3$, $e_{\min}=0.97$, $e_{\max}=1.30$.
&
8
&
$t_r=6$, same Canny-RNA rule.
\\

Bottleneck
&
\cmark
&
High--low--high bottleneck sampling.
&
4
&
$1024^2\!\rightarrow\!512^2\!\rightarrow\!1024^2$,
transition steps $(1,3)$.
&
8
&
$1024^2\!\rightarrow\!512^2\!\rightarrow\!1024^2$,
transition steps $(1,6)$.
\\

RaLu
&
\cmark
&
Region-adaptive lifting with timestep matching.
&
4
&
\makecell[l]{
$N=(2,1,1)$, $e=(0.092,0.233,1)$,\\
transition ratio $=0.30$.
}
&
--
&
--
\\

Speed
&
\xmark: DCT
&
Spectral progressive-resolution sampling.
&
4
&
DCT expansion, $t_r=3$, shift $0.97$.
&
8
&
DCT expansion, $t_r=6$.
\\

MrFlow
&
\xmark: Pixel SR
&
Pixel-space SR, VAE re-encoding, and one-step HR denoising.
&
$4+1$
&
\makecell[l]{
$(N_{\mathrm{LR}},N_{\mathrm{ref}})=(4,1)$,\\
Real-ESRGAN $\times2$.
}
&
$8+1$
&
\makecell[l]{
$(N_{\mathrm{LR}},N_{\mathrm{ref}})=(8,1)$,\\
Real-ESRGAN $\times2$.
}
\\

\bottomrule
\end{tabular}
\end{adjustbox}
\end{table*}
We report the implementation details required to reproduce the main
experiments. All evaluations generate $1024\times1024$ images using
FLUX.2-Klein-9b with 4 NFEs and Z-Image-Turbo with 8 NFEs. We first report the
additional architecture and optimization settings of SelfLift-rich, followed
by the exact configurations of the compared acceleration methods.

\subsection{SelfLift Hyperparameters}
\label{sec:supp_selflift_hyperparameters}
SelfLift-zero is training-free and uses the correction parameters reported in
the main paper. SelfLift-rich additionally employs a compact SwinIR-style
latent lifter and lightweight LoRA adaptation through On-Policy Self Recovery.
Table~\ref{tab:selflift_hyperparameters} lists the architecture and
optimization details omitted from the main text.

\subsection{Baseline Implementation Details}
\label{sec:supp_baseline_details}
We use the official implementations or released configurations of all
compared methods without altering their core acceleration mechanisms. Unless
defined otherwise by the original method, spatial baselines use a
$512\times512$ low-resolution stage and transition to $1024\times1024$ at
$t_r=3$ on FLUX.2-Klein-9b and $t_r=6$ on Z-Image-Turbo. MrFlow retains its
original pixel-space super-resolution pipeline. The shared mechanism and
model-specific settings of each baseline are summarized in
Table~\ref{tab:baseline_implementation_details}.
\begin{table}[t]
\centering
\caption{
Additional architecture, optimization, and training-cost details of
SelfLift-rich.
}
\label{tab:selflift_hyperparameters}

\fontsize{7.0pt}{7.4pt}\selectfont
\setlength{\tabcolsep}{3.2pt}
\renewcommand{\arraystretch}{1.00}
\begin{tabular}{
@{}p{2.20cm}
p{3.00cm}
>{\centering\arraybackslash}p{2.00cm}@{}
}
\toprule
\multicolumn{1}{c}{
  {\fontsize{5.2pt}{6.6pt}\selectfont\textbf{Category}}
}
&
\multicolumn{1}{c}{
  {\fontsize{5.2pt}{6.6pt}\selectfont\textbf{Parameter}}
}
&
\multicolumn{1}{c}{
  {\fontsize{5.2pt}{6.6pt}\selectfont\textbf{Value}}
}
\\
\midrule

\multirow{7}{*}{
\makecell[l]{\textbf{Latent Lifter}}
}
&
Input channels / scale
&
$128$ / $\times2$
\\

&
Embedding dimension
&
$192$
\\

&
Transformer depths
&
$(6,6,6,6)$
\\

&
Attention heads
&
$(8,8,8,8)$
\\

&
Window size / MLP ratio
&
$8$ / $2.5$
\\

&
Drop-path rate
&
$0.15$
\\

\midrule

\multirow{7}{*}{
\makecell[l]{\textbf{On-Policy}\\\textbf{Self Recovery}}
}
&
LoRA rank / alpha
&
$64$ / $128$
\\

&
Student learning rate
&
$2\times10^{-5}$
\\

&
EMA decay
&
$0.9999$
\\

&
Optimizer / betas
&
AdamW / $(0.9,0.999)$
\\

&
Weight decay
&
$0$
\\

&
Gradient clipping
&
$1.0$
\\

&
Max iterations
&
$3000$
\\

\midrule

\multirow{3}{*}{
\textbf{Training}
}
&
Batch size
&
$1$ per device
\\

&
Gradient accumulation
&
$1$
\\

&
Mixed precision
&
bf16
\\

\midrule

\multirow{3}{*}{
\makecell[l]{\textbf{Training Cost}}
}
&
Hardware
&
$8\times$ GPU 80GB
\\

&
Latent lifter
&
$5.56$ GPU-h
\\

&
LoRA adapter
&
$56.8$ GPU-h
\\

\bottomrule
\end{tabular}
\end{table}
\subsection{Evaluation Protocol and Metrics}
\label{sec:supp_evaluation_metrics}

All methods follow a unified protocol. GenEval uses its 300 official prompts,
while other metrics use a predefined balanced set of 300 prompts sampled
uniformly from DrawBench, the Pick-a-Pic test set, and the visual text-rendering
test split. Using shared seeds $\{0,42,128\}$, each main-table score averages
900 paired generations per method and backbone.
% , covering general,
% preference-oriented, and text-rich scenarios while reducing sampling variance.

\begin{itemize}
    \item \textbf{Inference efficiency:}
    End-to-end latency is measured on a single high-performance GPU using 10 warm-up
    runs followed by the average of 50 generations. Speedup is computed relative
    to Base (50), while latency reduction is measured against Base (4)/(8).
    
    \item \textbf{Overall generation quality:}
    We use ImageReward to assess the overall quality of generated images while
    accounting for their consistency with the input prompts.

    \item \textbf{Image quality and human preference:}
    PickScore evaluates preference-aligned visual quality, while AestheticScore
    measures the aesthetic appeal of the generated images.

    \item \textbf{Text alignment and instruction following:}
    CLIPScore measures global text--image semantic alignment. GenEval further
    evaluates compositional instruction following, including object counting,
    attribute binding, spatial relations, and color attribution.

    \item \textbf{Transition-quality ablation:}
    In the resolution-transition ablations, we additionally report CLIP-IQA as
    a sharpness-sensitive perceptual quality metric. It serves as a proxy for
    blur and visual-quality degradation introduced by different transition
    strategies.
\end{itemize}

\subsection{Runtime Breakdown}
\label{sec:supp_runtime_breakdown}

We further decompose the inference latency of FLUX.2-Klein-9b to identify
the sources of SelfLift's efficiency gains. All directly profiled results
follow the main evaluation protocol, using 10 warm-up generations followed
by 50 timed samples on a single GPU. Calculations use the unrounded
measurements, ensuring that the accumulated stage costs exactly recover the
corresponding end-to-end latency.
\par\medskip
\noindent
\begin{minipage}{\columnwidth}
\centering
\includegraphics[width=0.96\columnwidth]{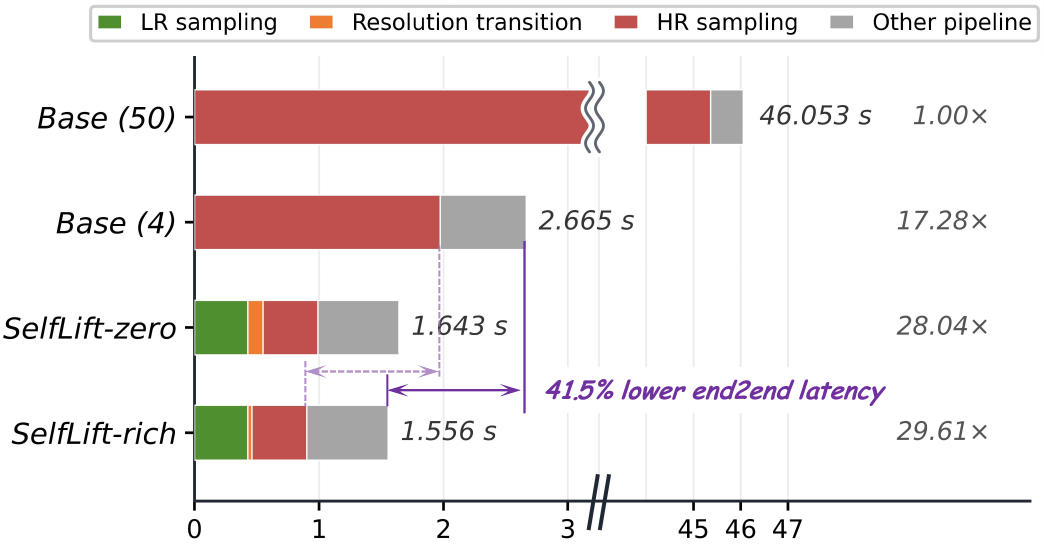}
\captionof{figure}{
Stage-wise runtime breakdown on FLUX.2-Klein-9b.
SelfLift-rich reduces end-to-end latency by $41.5\%$ relative to Base (4),
while its $3$-LR+$1$-HR schedule reduces denoising latency by $56.0\%$.
}
\label{fig:flux_runtime_breakdown}
\end{minipage}
\par\medskip
\par\medskip
\noindent
\begin{minipage}{0.96\columnwidth}
\centering
\captionof{table}{
Stage-wise runtime breakdown on FLUX.2-Klein-9b.
Times are reported in seconds per image. Totals are computed from
full-precision measurements, while displayed values are rounded to three
decimals.
}
\label{tab:flux_runtime_breakdown}

\footnotesize
\setlength{\tabcolsep}{1.5pt}
\renewcommand{\arraystretch}{0.95}

\begin{tabular}{
@{}
>{\raggedright\arraybackslash}p{2.10cm}
>{\centering\arraybackslash}p{1.65cm}
>{\centering\arraybackslash}p{1.15cm}
>{\centering\arraybackslash}p{1.15cm}
>{\centering\arraybackslash}p{1.15cm}
@{}
}
\toprule

\textbf{Component}
&
\textbf{\makecell{Base\\(50)}}
&
\textbf{\makecell{Base\\(4)}}
&
\textbf{\makecell{SelfLift\\ \emph{zero}}}
&
\textbf{\makecell{SelfLift\\ \emph{rich}}}
\\
\midrule

Sampling allocation
&
50$\times$2 HR
&
4 HR
&
3 L+1 H
&
3 L+1 H
\\

LR sampling
&
--
&
--
&
0.427
&
0.427
\\

HR sampling
&
45.362
&
1.973
&
0.441
&
0.441
\\

Resolution transition
&
--
&
--
&
0.123
&
0.033
\\

Other pipeline
&
0.691
&
0.692
&
0.652
&
0.655
\\

\midrule

\textbf{Total latency}
&
\textbf{46.053}
&
\textbf{2.665}
&
\textbf{1.643}
&
\textbf{1.556}
\\

Speedup
&
1.00$\times$
&
17.28$\times$
&
28.04$\times$
&
29.61$\times$
\\

\bottomrule
\end{tabular}
\end{minipage}
\par\medskip

\section{Extended Experiments}
\label{sec:extend-exp}
\begin{table*}[t]
\centering
\caption{
\textbf{Complete transition-strategy ablation corresponding to Fig.~5 of the main
paper.} Bold values denote the SelfLift-zero operating points adopted in the
main experiments: $t_r=3$ for FLUX.2-Klein-9b and $t_r=6$ for
Z-Image-Turbo.
}
\label{tab:full_transition_ablation}

\newcommand{\tabnumfont}{\fontsize{8.5pt}{8.9pt}\selectfont}
\setlength{\tabcolsep}{3.0pt}
\renewcommand{\arraystretch}{0.95}

\begin{adjustbox}{max width=0.98\textwidth}
\begin{tabular}{
@{}
>{\centering\arraybackslash}m{2.85cm}|
>{\raggedright\arraybackslash}m{2.55cm}
*{3}{>{\tabnumfont\centering\arraybackslash}c}
*{7}{>{\tabnumfont\centering\arraybackslash}c}
@{}
}
\toprule

\multirow{2}{*}{\textbf{Metric}}
&
\multirow{2}{*}{\textbf{\makecell[c]{Transition\\strategy}}}
&
\multicolumn{3}{c}{
    \footnotesize\textbf{FLUX.2-Klein-9b (4 NFEs)}
}
&
\multicolumn{7}{c}{
    \footnotesize\textbf{Z-Image-Turbo (8 NFEs)}
}
\\

\cmidrule(lr){3-5}
\cmidrule(lr){6-12}

&
&
\multicolumn{1}{c}{\footnotesize\textbf{$t_r=1$}}
&
\multicolumn{1}{c}{\footnotesize\textbf{$t_r=2$}}
&
\multicolumn{1}{c}{\footnotesize\textbf{$t_r=3$}}
&
\multicolumn{1}{c}{\footnotesize\textbf{$t_r=1$}}
&
\multicolumn{1}{c}{\footnotesize\textbf{$t_r=2$}}
&
\multicolumn{1}{c}{\footnotesize\textbf{$t_r=3$}}
&
\multicolumn{1}{c}{\footnotesize\textbf{$t_r=4$}}
&
\multicolumn{1}{c}{\footnotesize\textbf{$t_r=5$}}
&
\multicolumn{1}{c}{\footnotesize\textbf{$t_r=6$}}
&
\multicolumn{1}{c}{\footnotesize\textbf{$t_r=7$}}
\\

\midrule

\multirow{4}{*}{ImageReward$\uparrow$}
&
Latent branch
&
1.2852
&
1.3059
&
1.2778
&
1.0419
&
1.0538
&
1.0624
&
1.0445
&
1.0110
&
0.9594
&
0.8641
\\

&
Pixel branch
&
1.2004
&
1.2934
&
1.2878
&
1.0398
&
1.0287
&
1.0246
&
1.0334
&
1.0366
&
1.0382
&
1.0386
\\

&
Simple mix
&
1.2414
&
1.3124
&
1.2803
&
1.0236
&
1.0404
&
1.0431
&
1.0355
&
1.0413
&
1.0114
&
0.9864
\\

\rowcolor{red!5}
\cellcolor{white}
&
SelfLift-zero
&
1.2674
&
1.3166
&
\textbf{1.3013}
&
1.0281
&
1.0420
&
1.0544
&
1.0529
&
1.0582
&
\textbf{1.0354}
&
1.0204
\\

\midrule

\multirow{4}{*}{CLIP-IQA$\uparrow$}
&
Latent branch
&
0.8588
&
0.8721
&
0.8650
&
0.8197
&
0.8458
&
0.8534
&
0.8645
&
0.8600
&
0.8220
&
0.6606
\\

&
Pixel branch
&
0.7911
&
0.7815
&
0.7686
&
0.8021
&
0.7590
&
0.7683
&
0.7612
&
0.7525
&
0.7383
&
0.7162
\\

&
Simple mix
&
0.8393
&
0.8394
&
0.8220
&
0.7682
&
0.7945
&
0.8021
&
0.8016
&
0.7992
&
0.7732
&
0.5933
\\

\rowcolor{red!5}
\cellcolor{white}
&
SelfLift-zero
&
0.8577
&
0.8636
&
\textbf{0.8470}
&
0.8102
&
0.8336
&
0.8374
&
0.8379
&
0.8280
&
\textbf{0.7994}
&
0.7130
\\

\bottomrule
\end{tabular}
\end{adjustbox}
\end{table*}
\begin{table*}[t]
\centering
\caption{
\textbf{Complete ablation of On-Policy Self Recovery on FLUX.2-Klein-9b.}
All learned variants use the same transition step $t_r=3$.
Bold denotes the full SelfLift-rich model with
$\lambda_{\mathrm{dyn}}=80$, while underline marks the strongest
non-default variant for each quality metric.
}
\label{tab:full_rich_ablation}

\newcommand{\opsrtabnumfont}{\fontsize{8.5pt}{8.9pt}\selectfont}
\setlength{\tabcolsep}{3.0pt}
\renewcommand{\arraystretch}{0.9}

\begin{adjustbox}{max width=0.98\textwidth}
\begin{tabular}{
@{}
>{\raggedright\arraybackslash}m{4.30cm}|
>{\opsrtabnumfont\centering\arraybackslash}m{0.75cm}|
*{7}{>{\opsrtabnumfont\centering\arraybackslash}c}
@{}
}
\toprule

\multirow{2}{*}{\textbf{Variant}}
&
\multirow{2}{*}{$\boldsymbol{\lambda_{\mathrm{dyn}}}$}
&
\multicolumn{2}{c}{\footnotesize\textbf{Acceleration}}
&
\multicolumn{1}{c}{\footnotesize\textbf{Overall}}
&
\multicolumn{2}{c}{\footnotesize\textbf{Image Quality}}
&
\multicolumn{2}{c}{\footnotesize\textbf{Text Alignment}}
\\

\cmidrule(lr){3-4}
\cmidrule(lr){5-5}
\cmidrule(lr){6-7}
\cmidrule(lr){8-9}

&
&
\multicolumn{1}{c}{
    \footnotesize\textbf{Latency (s)$\downarrow$}
}
&
\multicolumn{1}{c}{
    \footnotesize\textbf{Speedup$\uparrow$}
}
&
\multicolumn{1}{c}{
    \footnotesize\textbf{ImageReward$\uparrow$}
}
&
\multicolumn{1}{c}{
    \footnotesize\textbf{PickScore$\uparrow$}
}
&
\multicolumn{1}{c}{
    \footnotesize\textbf{AestheticScore$\uparrow$}
}
&
\multicolumn{1}{c}{
    \footnotesize\textbf{CLIP$\uparrow$}
}
&
\multicolumn{1}{c}{
    \footnotesize\textbf{GenEval$\uparrow$}
}
\\

\midrule

SelfLift-zero
&
--
&
1.643
&
28.04$\times$
&
1.3013
&
22.6326
&
5.6481
&
32.317
&
0.8528
\\

\midrule

$+$ Learned latent lifter $\mathcal G_{\phi}$
&
--
&
1.556
&
29.61$\times$
&
1.3031
&
22.6737
&
5.6447
&
32.305
&
0.8554
\\

$+$ Off-policy distillation
&
--
&
1.556
&
29.61$\times$
&
1.1741
&
22.4648
&
5.5380
&
31.315
&
0.8084
\\

$+$ OPSR w/o $\mathcal L_{\mathrm{dyn}}$
&
0
&
1.556
&
29.61$\times$
&
1.1813
&
22.6381
&
5.6380
&
31.813
&
0.8292
\\

\midrule

$+$ OPSR
&
40
&
1.556
&
29.61$\times$
&
\underline{1.3048}
&
\underline{22.6819}
&
\underline{5.6706}
&
32.247
&
\underline{0.8701}
\\

\rowcolor{red!5}
\textbf{SelfLift-rich}
&
$\mathbf{80}$
&
\textbf{1.556}
&
\textbf{29.61$\times$}
&
\textbf{1.3171}
&
\textbf{22.7036}
&
\textbf{5.6875}
&
\textbf{32.345}
&
\textbf{0.8858}
\\

$+$ OPSR
&
120
&
1.556
&
29.61$\times$
&
1.3038
&
22.6467
&
5.6532
&
\underline{32.319}
&
0.8571
\\

\bottomrule
\end{tabular}
\end{adjustbox}
\end{table*}
\subsection{Ablations of SelfLift-zero}
\label{sec:zero-ablation}

We further examine the two key factors governing SelfLift-zero:
the spatial extent and strength of artifact-aware correction, controlled by
$\rho$ and $[w_{\min},w_{\max}]$, and the resolution-transition timestep
$t_r$. The former determines how the pixel-VAE anchor corrects the directly
lifted latent, while the latter controls the trade-off between low-resolution
computation and the remaining high-resolution recovery budget.

\subsubsection{Correction Parameters}
\label{sec:zero-parameter-ablation}

SelfLift-zero uses $\rho$ to select artifact-prone locations and
$[w_{\min},w_{\max}]$ to control the correction magnitude.
Table~\ref{tab:zero-parameter-ablation} isolates their effects through
one-factor-at-a-time ablations: three neighboring $\rho$ values on each
backbone and two alternative weight ranges on FLUX.2-Klein-9b.
All settings follow the main protocol with $t_r=3$ and $t_r=6$,
respectively. Since these parameters do not change the sampling schedule or
the number of model evaluations, latency remains unchanged.

\par\medskip
\noindent
\begin{minipage}{\columnwidth}
    \centering

    \captionof{table}{
        \textbf{Parameter ablation of SelfLift-zero.} Main-paper settings are bold.
        Panel (a) reports ImageReward and CLIP-IQA on both backbones.
        Panel (b) reports the remaining FLUX.2-Klein-9b evaluation metrics.
    }
    \label{tab:zero-parameter-ablation}

    \footnotesize
    \setlength{\tabcolsep}{3.1pt}
    \renewcommand{\arraystretch}{1.0}

    \def\ablationtablewidth{0.96\linewidth}

    (a) Correction parameters and primary metrics\\[-1pt]

    \begin{tabular*}{\ablationtablewidth}{
        @{\extracolsep{\fill}}lccc|cc@{}
    }
        \toprule
        ID
        & $\rho$
        & $w_{\min}$
        & $w_{\max}$
        & ImageReward$\uparrow$
        & CLIP-IQA$\uparrow$ \\
        \midrule

        \multicolumn{6}{c}{
            FLUX.2-Klein-9b: selected-region ratio
        } \\
        F1
        & 0.3 & 0.5 & 1.0
        & 1.2948 & \textbf{0.8510} \\
        \textbf{F2}
        & \textbf{0.4}
        & \textbf{0.5}
        & \textbf{1.0}
        & \textbf{1.3013}
        & 0.8470 \\
        F3
        & 0.5 & 0.5 & 1.0
        & 1.2967 & 0.8390 \\

        \midrule
        \multicolumn{6}{c}{
            Z-Image-Turbo: selected-region ratio
        } \\
        Z1
        & 0.2 & 0.5 & 1.0
        & 1.0278 & \textbf{0.8020} \\
        \textbf{Z2}
        & \textbf{0.3}
        & \textbf{0.5}
        & \textbf{1.0}
        & \textbf{1.0354}
        & 0.7994 \\
        Z3
        & 0.4 & 0.5 & 1.0
        & 1.0306 & 0.7870 \\

        \midrule
        \multicolumn{6}{c}{
            FLUX.2-Klein-9b: correction weights (default: F2)
        } \\
        W1
        & 0.4 & 0.5 & 0.5
        & 1.2899 & \textbf{0.8500} \\
        W2
        & 0.4 & 0.8 & 1.0
        & 1.2938 & 0.8360 \\

        \bottomrule
    \end{tabular*}

    \vspace{10pt}

    (b) Remaining FLUX.2-Klein-9b metrics\\[-1pt]

    \begin{tabular*}{\ablationtablewidth}{
        @{\extracolsep{\fill}}lcccc@{}
    }
        \toprule
        ID
        & PickScore$\uparrow$
        & Aesthetic$\uparrow$
        & CLIP$\uparrow$
        & GenEval$\uparrow$ \\
        \midrule

        F1
        & 22.612
        & \textbf{5.654}
        & 32.300
        & 0.848 \\
        \textbf{F2}
        & \textbf{22.633}
        & 5.648
        & \textbf{32.317}
        & \textbf{0.853} \\
        F3
        & 22.621
        & 5.639
        & 32.306
        & 0.850 \\

        \midrule
        W1
        & 22.602
        & 5.651
        & 32.291
        & 0.846 \\
        W2
        & 22.615
        & 5.635
        & 32.300
        & 0.849 \\

        \bottomrule
    \end{tabular*}
\end{minipage}
\par\medskip

Across both backbones, smaller $\rho$ confines correction to high-risk
regions, preserving sharpness but leaving more artifacts, whereas larger
$\rho$ improves stability at the cost of smoothing. We therefore use
$\rho=0.4$ for FLUX.2-Klein-9b and $\rho=0.3$ for Z-Image-Turbo.
The range $[w_{\min},w_{\max}]$ controls correction strength: $w_{\min}$
sets the correction floor, while $w_{\max}$ caps correction in the most
inconsistent regions. Increasing $w_{\min}$ suppresses transition artifacts
and produces more stable outputs, but may reduce CLIP-IQA through
over-smoothing. The default $[0.5,1.0]$ balances stability and sharpness and
can be adjusted with negligible overhead across model families. For
aggressive transitions or transition-sensitive tasks such as text rendering,
a larger $w_{\min}$ is recommended to obtain stronger pixel-VAE correction
and more reliable generation.

\subsubsection{Resolution-Transition Timestep}
\label{sec:transition-timestep-ablation}

Table~\ref{tab:full_transition_ablation} reports the complete numerical
results corresponding to Fig.~5 of the main paper. We evaluate the latent
branch, pixel branch, simple mix, and SelfLift-zero at every valid
resolution-transition timestep on both backbones.
\begin{figure}[t]
\centering
\includegraphics[width=0.90\columnwidth]
{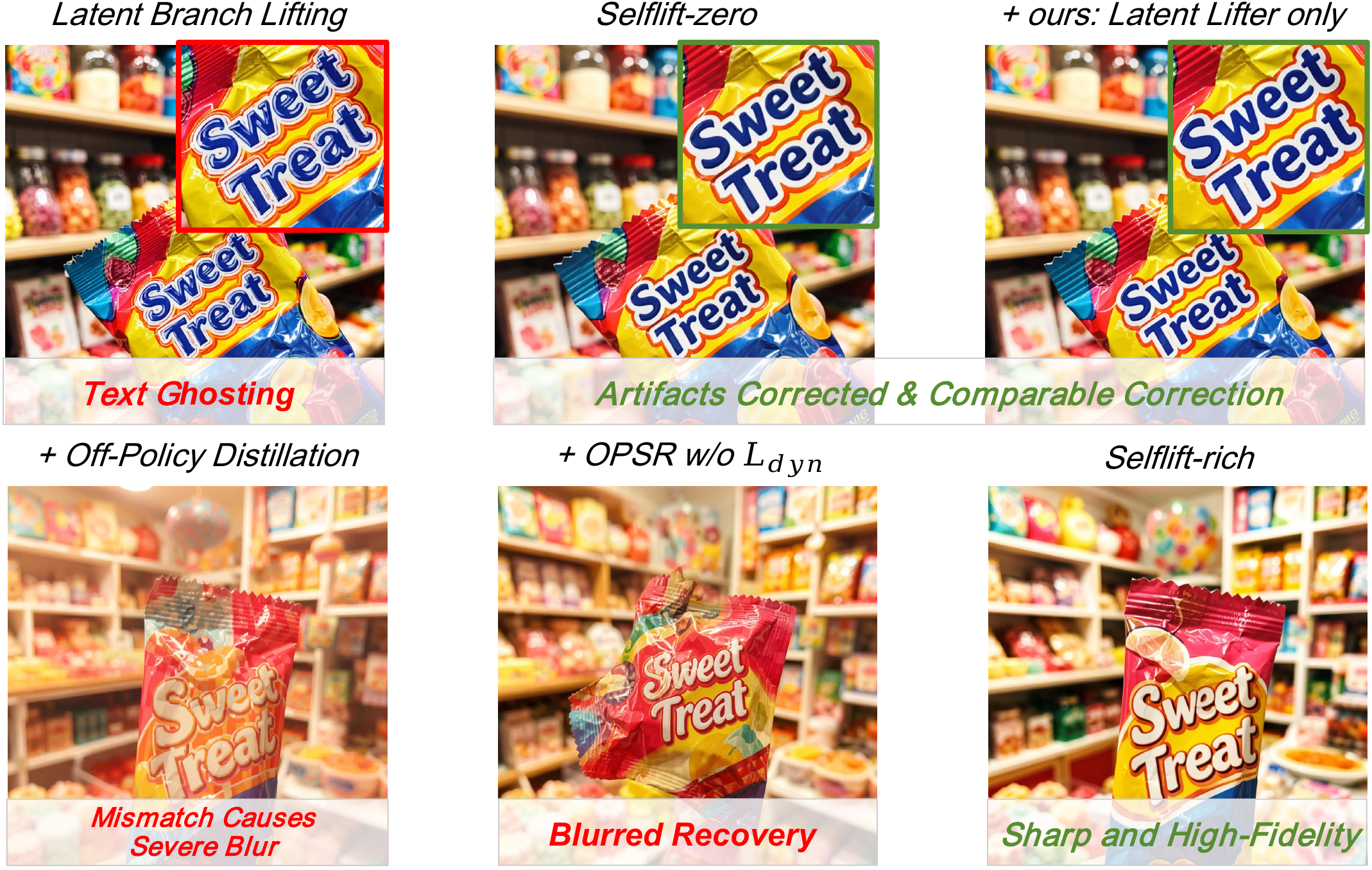}
\caption{
\textbf{Qualitative ablation of On-Policy Self Recovery.}
Top: SelfLift-zero and the learned lifter correct direct-lifting ghosting,
with the latter bypassing the pixel--VAE path. Bottom: off-policy
distillation produces blur, removing $\mathcal L_{\mathrm{dyn}}$ weakens
recovery, and SelfLift-rich remains sharp.
    }
\label{fig:supp_ablation2}
\end{figure}
\subsection{Ablation of On-Policy Self Recovery}
\label{sec:opsr-ablation}
\begin{figure*}[t]
    \centering
    \includegraphics[width=0.98\textwidth]
    {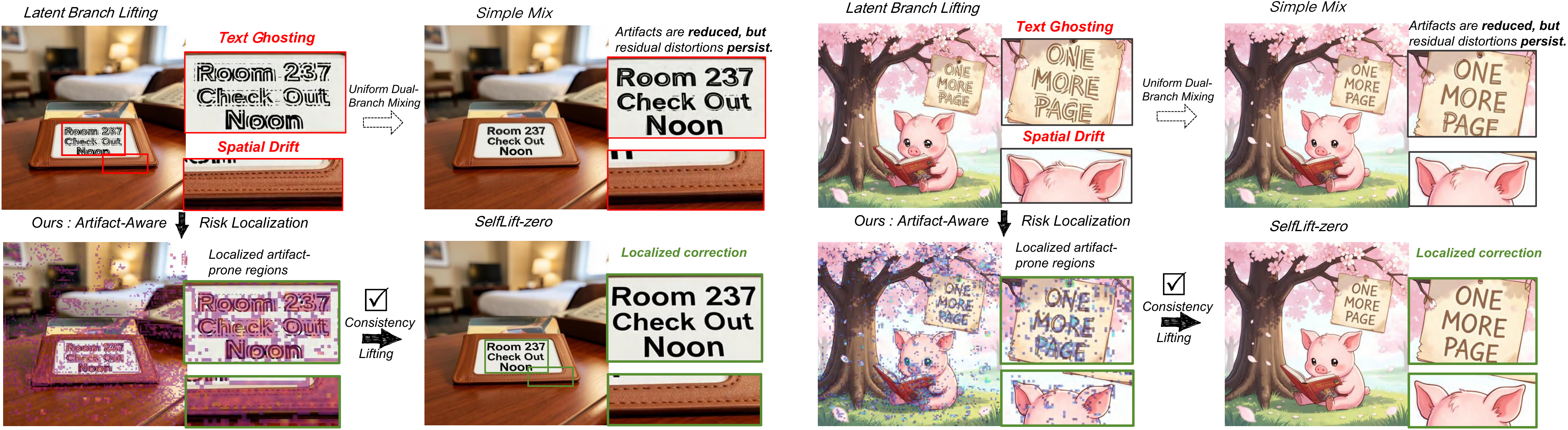}
    \caption{
    \textbf{Additional validation of artifact localization.}
    Predicted high-risk regions align with visible ghosting and spatial drift.
    Selective correction removes these localized artifacts, whereas uniform
    mixing leaves residual distortions. This extends Fig.~6 of the main paper.
    }
    \label{fig:supp_artifact_localization}
\end{figure*}
Increasing $t_r$ preserves more computation at low resolution and therefore
provides greater acceleration, but leaves fewer high-resolution steps for
recovering transition errors. Direct latent lifting retains relatively high
sharpness at moderate transition timesteps, yet its prompt-aligned quality
degrades sharply under aggressive late transitions, particularly on
Z-Image-Turbo. Pixel-VAE re-encoding remains more stable at late timesteps
but consistently exhibits lower CLIP-IQA because of over-smoothing.
Simple mixing merely interpolates between these two failure modes.
SelfLift-zero instead maintains a favorable sharpness--alignment trade-off at
the selected operating points, $t_r=3$ for FLUX.2-Klein-9b and $t_r=6$ for
Z-Image-Turbo.

To complement the main-paper OPSR ablation, we provide qualitative evidence
and a controlled study on FLUX.2-Klein-9b. We isolate three additions to
SelfLift-zero: a learned lifter replacing the pixel--VAE round trip,
trajectory-aligned on-policy supervision, and dynamics preservation. All
learned variants use $t_r=3$.

Figure~\ref{fig:supp_ablation2} links these components to their failure
modes. The learned lifter removes text ghosting and matches SelfLift-zero
without requiring pixel--VAE inference. Off-policy supervision causes severe
blur, removing dynamics preservation yields incomplete recovery, and the
full model produces sharp results.

Table~\ref{tab:full_rich_ablation} reports all metrics and a dynamics-weight
sweep. The lifter reduces transition latency from $1.642$ to $1.556$ seconds
without degrading quality. Off-policy distillation degrades all metrics,
while on-policy supervision remains insufficient without
$\mathcal L_{\mathrm{dyn}}$. With other settings fixed,
$\lambda_{\mathrm{dyn}}=80$ performs best; $40$ under-regularizes dynamics,
whereas $120$ limits adaptation and approaches SelfLift-zero. These results
confirm the complementarity of on-policy recovery and dynamics preservation.

\subsection{Additional Validation of Artifact Localization}
\label{sec:additional-artifact-localization}

Figure~\ref{fig:supp_artifact_localization} further validates Fig.~6 on two
examples. Localized correction
removes these artifacts while preserving reliable content, whereas uniform
mixing leaves residual distortions, confirming that the consistency residual
captures artifact-sensitive transition errors.

\subsection{Additional Qualitative Results}
\label{sec:supp-qualitative}

Figures~\ref{fig:supp_flux_qualitative} and
\ref{fig:supp_zimage_qualitative} extend the main-paper comparison with
additional prompts on FLUX.2-Klein-9b and Z-Image-Turbo, further demonstrating
that SelfLift consistently suppresses transition artifacts while preserving
text accuracy, local structure, and fine-grained visual details across model
families.

\section{Applicability and practical adaptation.}
SelfLift-zero is a training-free, plug-and-play transition operator applicable across model families without modifying the denoiser or sampling schedule. Its effectiveness depends on the predicted clean sample used to construct both the direct latent and pixel-VAE branches.  At early timesteps, the clean estimate
is still immature, making the pixel branch and the resulting correction
signal less reliable. Transitioning too early may preserve residual noise or incomplete structure, producing an under-denoised appearance while requiring more high-resolution steps and reducing acceleration.
SelfLift-zero is therefore better suited to aggressive late-transition
configurations, where the clean estimate is sufficiently developed,
pixel-VAE guidance becomes reliable, and the errors accumulated by direct
latent lifting are more pronounced. As transition correction is decoupled from timing, SelfLift-zero complements
schedulers such as Speed, especially for flexible multi-step models.

The same plug-and-play principle can extend to video generation when the backbone reliably supports the chosen resolutions. In our
preliminary exploration with Wan2.1-T2V, $2\times$ spatial downsampling of 480p inputs altered the token-sequence distribution and destabilized scene structure, indicating poor generalization to unseen sequence lengths. Progressive-resolution video inference should therefore remain within the model's native multi-resolution regime, where SelfLift can reliably bridge resolution stages without additional training or external restoration models.

% \begin{figure*}[!t]
%     \centering
%     \includegraphics[width=0.98\textwidth]
%     {Figures/supp_exp_show_v1.pdf}
%     \caption{
%     \textbf{Additional qualitative comparisons on FLUX.2-Klein-9b-4NFEs.}
%     Columns show Original, Vanilla, RaLu, Speed, MrFlow, SelfLift-zero, and
%     SelfLift-rich. Red boxes show the enlarged regions.
%     }
%     \label{fig:supp_flux_qualitative}
% \end{figure*}

% \begin{figure*}[!t]
%     \centering
%     \includegraphics[width=0.98\textwidth]
%     {Figures/supp_exp_show2_v1.pdf}
%     \caption{
%     \textbf{Additional qualitative comparisons on Z-Image-Turbo-8NFEs.}
%     Columns show Original, Vanilla, Bottleneck, Speed, MrFlow, SelfLift-zero,
%     and SelfLift-rich. Red boxes show the enlarged regions.
%     }
%     \label{fig:supp_zimage_qualitative}
% \end{figure*}

\end{document}